\documentclass{article}

\usepackage{microtype}
\usepackage{graphicx}
\usepackage{subfigure}

\usepackage{amsmath}
\usepackage{amsthm}
\usepackage{amsfonts}
\usepackage{todonotes}
\usepackage{mathtools}

\usepackage{booktabs} 

\usepackage{hyperref}

\usepackage[accepted]{icml2020}

\newcommand{\Exs}{\ensuremath{\mathbb{E}}}

\newtheorem{theo}{Theorem}[section]

\newtheorem{lem}{Lemma}[section]
\newtheorem{prop}{Proposition}[section]
\newtheorem{cor}{Corollary}[section]

\newtheorem{nota}{Notation}[section]
\newtheorem{de}{Definition}[section]
\newtheorem{exa}{Example}[section]
\newtheorem{as}{Assumption}[section]
\newtheorem{alg}{Algorithm}[section]

\newcommand{\btheo}{\begin{theo}}
\newcommand{\bde}{\begin{de}}
\newcommand{\ble}{\begin{lem}}
\newcommand{\bpr}{\begin{prop}}
\newcommand{\bno}{\begin{nota}}
\newcommand{\bex}{\begin{exa}}
\newcommand{\bcor}{\begin{cor}}
\newcommand{\spro}{\begin{proof}}
\newcommand{\bas}{\begin{as}}
\newcommand{\balg}{\begin{alg}}

\newcommand{\etheo}{\end{theo}}
\newcommand{\ede}{\end{de}}
\newcommand{\ele}{\end{lem}}
\newcommand{\epr}{\end{prop}}
\newcommand{\eno}{\end{nota}}
\newcommand{\eex}{\end{exa}}
\newcommand{\ecor}{\end{cor}}
\newcommand{\fpro}{\end{proof}}
\newcommand{\eas}{\end{as}}
\newcommand{\ealg}{\end{alg}}

\newtheorem{theos}{Theorem}
\newtheorem{props}{Proposition}
\newtheorem{lems}{Lemma}
\newtheorem{cors}{Corollary}

\newtheorem{exas}{Example}
\newtheorem{algs}{Algorithm}
\newtheorem{asss}{Assumption}
\newtheorem{defns}{Definition}

\newcommand{\btheos}{\begin{theos}}
\newcommand{\etheos}{\end{theos}}
\newcommand{\bprops}{\begin{props}}
\newcommand{\eprops}{\end{props}}
\newcommand{\bdes}{\begin{defns}}
\newcommand{\edes}{\end{defns}}
\newcommand{\blems}{\begin{lems}}
\newcommand{\elems}{\end{lems}}
\newcommand{\bcors}{\begin{cors}}
\newcommand{\ecors}{\end{cors}}
\newcommand{\bexs}{\begin{exas}}
\newcommand{\eexs}{\end{exas}}
\newcommand{\balgs}{\begin{algs}}
\newcommand{\ealgs}{\end{algs}}
\newcommand{\bass}{\begin{asss}}
\newcommand{\eass}{\end{asss}}

\DeclareMathOperator{\tr}{tr} 

\begin{document}

\twocolumn[
\icmltitle{Distributed Averaging Methods for Randomized Second Order Optimization}








\begin{icmlauthorlist}
\icmlauthor{Burak Bartan}{su}
\icmlauthor{Mert Pilanci}{su}
\end{icmlauthorlist}

\icmlaffiliation{su}{Department of Electrical Engineering, Stanford University, California, USA}

\icmlcorrespondingauthor{Burak Bartan}{bbartan@stanford.edu}
\icmlcorrespondingauthor{Mert Pilanci}{pilanci@stanford.edu}

\icmlkeywords{Machine Learning, ICML}

\vskip 0.3in
]



\printAffiliationsAndNotice{}  

\begin{abstract}
We consider distributed optimization problems where forming the Hessian is computationally challenging and communication is a significant bottleneck. We develop unbiased parameter averaging methods for randomized second order optimization that employ sampling and sketching of the Hessian. Existing works do not take the bias of the estimators into consideration, which limits their application to massively parallel computation. We provide closed-form formulas for regularization parameters and step sizes that provably minimize the bias for sketched Newton directions. We also extend the framework of second order averaging methods to introduce an \textit{unbiased} distributed optimization framework for heterogeneous computing systems with varying worker resources. Additionally, we demonstrate the implications of our theoretical findings via large scale experiments performed on a serverless computing platform.

\end{abstract}
\section{Introduction} \label{sec:intro}

We consider distributed averaging in a variety of randomized second order methods including Newton Sketch, iterative Hessian sketch (IHS), and also in \textit{direct} (i.e. non-iterative) methods for solving regularized least squares problems. Averaging sketched solutions was proposed in the literature in certain restrictive settings \cite{wang2018distributed}. The presence of a regularization term requires additional caution, as na\"ive averaging may lead to biased estimators of the solution. Although this is often overlooked in the literature, we show that one can re-calibrate the regularization coefficient to obtain unbiased estimators. We show that having unbiased estimators leads to better performance without imposing any additional computational cost.

Our bias correction results have additional desirable properties for distributed computing systems. For heterogeneous distributed computing environments, where workers have varying computing capabilities, it might be advantageous for each worker to solve a problem of a different size \cite{avestimehr2017heterogenous_coded}. We provide formulas that specify the regularization parameter as a function of the problem size for each worker to obtain an unbiased estimator of the optimal solution.

Serverless computing is a relatively new technology that offers computing on the cloud without requiring any server management from end users. Workers in serverless computing usually have very limited resources and lifetime, but also are very scalable. The algorithms we study in this paper are particularly suitable for serverless computing platforms, since the algorithms do not require peer-to-peer communication among worker nodes,  and have low memory and compute requirements per node. In numerical simulations, we have evaluated our methods on the serverless computing platform AWS Lambda.

\subsection{Previous Work and Our Contributions}

In this work, we study averaging for randomized second order methods for Least Squares problems, as well as a more general class of convex optimization problems.

Random projections are a popular way of performing randomized dimensionality reduction, which are widely used in many computational and learning problems ~\citep{vempala2005random, mahoney2011randomized, woodruff2014sketching, drineas2016randnla}. Many works have studied randomized sketching methods for least squares and optimization problems \citep{avron2010blendenpik, rokhlin2009randomized, drineas2011faster, pilanci2015randomized,pilanci2017newton,mahoney2018averaging,mahoney2018giant}.

Our results on least squares regression improve on the results in \cite{mahoney2018averaging} for averaging multiple sketched solutions. In particular, in \cite{mahoney2018averaging}, the sketched sub-problems use the same regularization parameter as the original problem, which leads to biased solutions. We analyze the bias of the averaged solution, and provide explicit formulas for selecting the regularization parameter of the sketched sub-problems to achieve unbiasedness. In addition, we analyze the convergence rate of the distributed version of the iterative Hessian sketch algorithm which was introduced in \cite{PilWai14b}.

One of the main contributions of this work is developing bias correction for averaging sketched solutions. Namely, the setting considered in \cite{mahoney2018giant} is based on a distributed second order optimization method that involves averaging approximate update directions. However, in that work, the bias was not taken into account which degrades accuracy and limits applicability to massively parallel computation.
We additionally provide results for the unregularized case, which corresponds to the distributed version of the Newton sketch algorithm introduced in \cite{pilanci2017newton}

\subsection{Paper Organization}
In Section \ref{sec:prelim}, we describe the problem setup and notation and discuss different types of sketching matrices we consider. Section \ref{sec:ihs} presents Theorem \ref{thm_IHS_error_decay} and Corollary \ref{ihs_main_corollary} which establish the error decay and convergence properties of the distributed iterative Hessian sketch algorithm given in Algorithm \ref{alg:dist_ihs}. Section \ref{sec:regularized_least_squares} presents Theorem \ref{opt_lambda_2_newton_LS}, which characterizes bias conditions for the averaged ridge regression estimator. Sections \ref{sec:newton_sketch} and \ref{sec:regularized_newton} provide an analysis of the bias of estimators for the averaged Newton sketch update directions with no regularization (Theorem \ref{thm:dist_newton_sketch}) and regularization (Theorem \ref{opt_lambda_2_newton}), respectively. Section \ref{sec:example_problems} discusses two optimization problems where our results can be applied. In Section \ref{sec:numerical_results}, we present our numerical results.

The proofs of the theorems and lemmas are provided in the supplementary material.
\section{Preliminaries} \label{sec:prelim}

\subsection{Problem Setup and Notation}
We consider a distributed computing model where we have $q$ worker nodes and a single central node, i.e., master node. The workers may only be allowed to communicate with the master node. The master collects the outputs of the worker nodes and returns the averaged result. For iterative algorithms, this step serves as a synchronization point.

Throughout the text, we provide exact formulas for the bias and variance of sketched solutions. All the expectations in the paper are with respect to the randomness over the sketching matrices, where no randomness assumptions are made for the data. 

Throughout the text, we use hats (e.g. $\hat{x}_k$) to denote the estimator for the $k$'th sketch and bars (e.g. $\bar{x}$) to denote the averaged estimator. We use $f(.)$ to denote the objective of whichever optimization problem is being considered at that point in the text.

We use $S \in \mathbb{R}^{m \times n}$ to denote random sketching matrices. For non-iterative distributed algorithms, we use $S_k \in \mathbb{R}^{m \times n}$ to refer to the sketching matrix used by worker $k$. For iterative algorithms, $S_{t,k} \in \mathbb{R}^{m \times n}$ is used to denote the sketching matrix used by worker $k$ in iteration $t$. We assume the sketching matrices are appropriately scaled to satisfy $\Exs[S_{t,k}^TS_{t,k}]=I_n$ and are independently drawn by each worker. We omit the subscripts in $S_{k}$ and $S_{t,k}$ for simplicity whenever it does not cause confusion.

For problems involving regularization, we use $\lambda_1$ for the regularization coefficient of the original problem, and $\lambda_2$ for the regularization coefficient of the sketched sub-problems.

\subsection{Sketching Matrices}
We consider various sketching matrices in this work including Gaussian sketch, uniform sampling, randomized Hadamard based sketch, Sparse Johnson-Lindenstrauss Transform (SJLT), and \textit{hybrid} sketch. We now briefly describe each of these sketching methods:

\begin{enumerate}
    \item \textit{Gaussian sketch:} Entries of $S \in \mathbb{R}^{m\times n}$ are i.i.d. and sampled from the Gaussian distribution. Sketching a matrix $A \in \mathbb{R}^{n\times d}$ using Gaussian sketch requires matrix multiplication $SA$ which has computational complexity equal to $\mathcal{O}(mnd)$.
    
    \item \textit{Randomized Hadamard based sketch:} The sketch matrix in this case can be represented as $S=PHD$ where $P \in \mathbb{R}^{m\times n}$ is for uniform sampling of $m$ rows out of $n$ rows, $H \in \mathbb{R}^{n\times n}$ is the Hadamard matrix, and $D \in \mathbb{R}^{n\times n}$ is a diagonal matrix with diagonal entries sampled randomly from the Rademacher distribution. Multiplication by $D$ to obtain $DA$ requires $\mathcal{O}(nd)$ scalar multiplications. Hadamard transform can be implemented as a fast transform with complexity $\mathcal{O}(n\log(n))$ per column, and a total complexity of $\mathcal{O}(nd\log(n))$ to sketch all $d$ columns of $DA$. We note that because $P$ reduces the row dimension down to $m$, it might be possible to devise a more efficient way to perform sketching with lower computational complexity.
    
    \item \textit{Uniform sampling:} Uniform sampling randomly selects $m$ rows out of the $n$ rows of $A$ where the probability of any row being selected is the same.
    
    \item \textit{Sparse Johnson-Lindenstrauss Transform (SJLT) \cite{nelson2013osnap}:} The sketching matrix for SJLT is a sparse matrix where each column has exactly $s$ nonzero entries and the columns are independently distributed. The nonzero entries are sampled from the Rademacher distribution. It takes $\mathcal{O}(snd/m)$ addition operations to sketch a data matrix using SJLT.
    
    \item \textit{Hybrid sketch:} The method that we refer to as hybrid sketch is a sequential application of two different sketching methods. In particular, it might be computationally feasible for worker nodes to sample as much data as possible ($m_2$ rows) and then reduce the dimension of the available data to the final sketch dimension $m$ using another sketch with better properties than uniform sampling such as Gaussian sketch or SJLT. For instance, hybrid sketch with uniform sampling followed by Gaussian sketch has computational complexity $\mathcal{O}(mm_2d)$.
\end{enumerate}

\section{Distributed Iterative Hessian Sketch} \label{sec:ihs}
In this section, we consider the well-known problem of unconstrained linear least squares which is stated as
\begin{align} \label{eq:standard_least_sq}
    x^* = \arg \min_x \frac{1}{2}\|Ax-b \|_2^2,
\end{align}
where $A \in \mathbb{R}^{n \times d}$ and $b \in \mathbb{R}^n$ are the problem data. Newton's method terminates in one step when applied to this problem since the Hessian is $A^TA$ and
\begin{align*}
    x_{t+1} = x_t - \mu (A^TA)^{-1}A^T(Ax_t-b).
\end{align*}
However, the computational cost of this direct solution is often prohibitive for large scale problems. Iterative Hessian sketch introduced in \cite{PilWai14b} employs a randomly sketched Hessian $A^TS_t^T S_t^T A$ as follows
\begin{align*}
    x_{t+1} = x_t - \mu (A^TS_t^TS_tA)^{-1}A^T(Ax_t-b),
\end{align*}
where $S_t$ corresponds to the sketching matrix at iteration $t$. Sketching reduces the row dimension of the data from $n$ to $m$ and hence computing an approximate Hessian $A^TS_t^TS_tA$ is computationally cheaper than the exact Hessian $A^TA$. Moreover, for regularized problems one can choose $m$ smaller than $d$ as we investigate in Section \ref{sec:regularized_least_squares}.

In a distributed computing setting, one can obtain more accurate update directions by averaging multiple trials, where each worker node computes an independent estimate of the update direction. These approximate update directions can be averaged at the master node and the following update takes place
\begin{align} \label{averaged_ihs_update}
    x_{t+1} = x_t - \mu \frac{1}{q} \sum_{k=1}^q (A^TS_{t,k}^TS_{t,k}A)^{-1}A^T(Ax_t-b).
\end{align}
Here $S_{t,k}$ is the sketching matrix for the $k$'th worker at iteration $t$. The details of distributed IHS algorithm are given in Algorithm \ref{alg:dist_ihs}. 
We note note that the above update can be replaced with an approximate solution. It might be computationally more efficient for worker nodes to obtain their approximate update directions using indirect methods such as conjugate gradient.

Note that workers communicate their approximate update directions and not the approximate Hessian matrix, which reduces the communication complexity from $\mathcal{O}(d^2)$ to $\mathcal{O}(d)$ for each worker per iteration.

\begin{algorithm}[tb]
  \caption{Distributed Iterative Hessian Sketch}
  \label{alg:dist_ihs}
\begin{algorithmic}
  \STATE {\bfseries Input:} Number of iterations $T$, step size $\mu$.
  \FOR{$t=1$ {\bfseries to} $T$}
    \FOR{workers $k=1$ {\bfseries to} $q$ in parallel}
        \STATE Sample $S_{t,k} \in \mathbb{R}^{m\times n}$.
        \STATE Sketch the data $S_{t,k}A$.
        \STATE Compute gradient $g_t = A^T(Ax_t-b)$.
        \STATE Solve $\hat{\Delta}_{t,k}=\arg\min_{\Delta} \frac{1}{2} \|S_{t,k} A \Delta \|_2^2 + g_t^T\Delta $ and send to master.
    \ENDFOR
    \STATE \textbf{Master:} Update $x_{t+1} = x_t + \mu \frac{1}{q}\sum_{k=1}^q \hat{\Delta}_{t,k}$ and send $x_{t+1}$ to workers.
  \ENDFOR
  \STATE return $x_T$
\end{algorithmic}
\end{algorithm}

We establish the convergence rate for Gaussian sketches in Theorem \ref{thm_IHS_error_decay}, which provides an exact result of the expected error.

\begin{de} \label{error_vec_def}
To quantify the approximation quality of the iterate $x_t \in \mathbb{R}^d$ with respect to the optimal solution $x^* \in \mathbb{R}^d$, we define the error as $e^A_t \coloneqq A(x_t-x^*)$ where $A \in \mathbb{R}^{n\times d}$ is the data matrix.
\end{de}
To state our result, we first introduce the following moments of the inverse Wishart distribution (see  Appendix).
\begin{align} \label{eq:theta_1_2_definitions}
        \theta_1 &\coloneqq \frac{m}{m-d-1}, \nonumber \\
     \theta_2 &\coloneqq \frac{m^2(m-1)}{(m-d)(m-d-1)(m-d-3)}.
\end{align}

\btheos[Expected Error Decay for Gaussian Sketches] \label{thm_IHS_error_decay}
In Algorithm \ref{alg:dist_ihs}, the expected squared norm of the error $e^A_{t}$, when we set $\mu = 1/\theta_1$ and $S_{t,k}$'s are i.i.d. Gaussian sketches evolves according to the following relation:
\begin{align*}
\Exs [\| e^A_{t+1} \|_2^2] = \frac{1}{q} \left( \frac{\theta_2}{\theta_1^2} - 1 \right) \|e^A_t\|_2^2.
\end{align*}
\etheos

The next corollary characterizes the number of iterations for Algorithm \ref{alg:dist_ihs} to achieve an error of $\epsilon$, and states that the number of iterations required for error $\epsilon$ scales with $\log(1/\epsilon)/\log(q)$.

\bcors \label{ihs_main_corollary}
Let $S_{t,k} \in \mathbb{R}^{m\times n}$ ($t=1,...,T$, $k=1,...,q$) be Gaussian sketching matrices. Then, Algorithm \ref{alg:dist_ihs} outputs $x_T$ that is $\epsilon$-accurate with respect to the initial error in expectation, that is, $\frac{\Exs [\|e^A_T\|_2^2}{\|Ax^*\|_2^2} = \epsilon$ where $T$ is given by
\begin{align*}
    T &= \frac{\log(1/\epsilon)}{\log(q)-\log\left(\frac{\theta_2}{\theta_1^2}-1\right)},
\end{align*}
where the overall required communication is $Tqd$ numbers, and the computational complexity per worker is 
\begin{align*}
    \mathcal{O}(Tmnd + Tmd^2 + Td^3).
\end{align*}
\ecors
\textit{Remark:} Provided that $m$ is at least $2d$, the term $\log\left(\frac{\theta_2}{\theta_1^2}-1\right)$ is negative. Hence, $T$ is upper-bounded by $\frac{\log(1/\epsilon)}{\log(q)}$.

\section{Averaging for Regularized Least Squares} \label{sec:regularized_least_squares}
The method described in this section is based on non-iterative averaging for solving the linear least squares problem with $\ell_2$ regularization, i.e., ridge regression, and is fully asynchronous.

We consider the problem given by
\begin{align}
    x^* = \arg\min_x \|Ax-b\|_2^2 + \lambda_1 \|x\|_2^2,
\end{align}
where $A \in \mathbb{R}^{n\times d}$, $b \in \mathbb{R}^n$ denote input data, and $\lambda_1 > 0$ is a regularization parameter.
Each worker applies sketching on $A$ and $b$ and obtains the estimate $\hat{x}_k$ given by
\begin{align}
    \hat{x}_k = \arg\min_x \|S_kAx-S_kb\|_2^2 + \lambda_2 \|x\|_2^2
\end{align}
for $k=1,...,q$,
and the averaged solution is computed by the master node as
\begin{align}
    \bar{x} = \frac{1}{q} \sum_{k=1}^q \hat{x}_k.
\end{align}
Note that we have $\lambda_1$ as the regularization coefficient of the original problem and $\lambda_2$ for the sketched sub-problems. If $\lambda_2$ is chosen to be equal to $\lambda_1$, then this scheme reduces to the framework given in the work of \cite{mahoney2018averaging} and we show in Theorem \ref{opt_lambda_2_newton_LS} that $\lambda_2 = \lambda_1$ leads to a biased estimator, which does not converge to the optimal solution.

We next introduce the following results on traces involving random Gaussian matrices which are instrumental in our result.

\blems [\cite{liu2019ridge}] \label{expectation_inverse_regularization}
For a Gaussian sketching matrix $S$, the following holds
\begin{align*}
    \lim_{n \rightarrow \infty} \Exs[\tr( (U^TS^TSU + \lambda_2 I)^{-1} )] = d \times \theta_3(d/m, \lambda_2),
\end{align*}
where $\theta_3(d/m, \lambda_2)$ is defined as $\theta_3(d/m, \lambda_2) = $
\begin{align*}
    = \frac{-\lambda_2+d/m-1 + \sqrt{(-\lambda_2+d/m-1)^2+4\lambda_2d/m}}{2\lambda_2 d/m}.
\end{align*}
\elems

\blems [] \label{expectation_inverse_regularization_2}
For a Gaussian sketching matrix $S$, the following holds
\begin{align*}
    \lim_{n \rightarrow \infty} \Exs[(U^TS^TSU + \lambda_2 I)^{-1}] = \theta_3(d/m, \lambda_2) I_d,
\end{align*}
where $\theta_3(d/m, \lambda_2)$ is as defined in Lemma \ref{expectation_inverse_regularization}.
\elems

\btheos[] \label{opt_lambda_2_newton_LS}
Given the thin SVD decomposition $A=U\Sigma V^T \in \mathbb{R}^{n\times d}$ and $n \geq d$, and assuming $A$ has full rank and has identical singular values (i.e., $\Sigma = \sigma I_d$), there is a value of $\lambda_2$ that yields a zero bias of the single-sketch estimator $\Exs [A(\hat{x}_k - x^*)]$ as $n$ goes to infinity if 
\begin{itemize}
    \item[(i)] $m > d$\,\, or
    \item[(ii)] $m \leq d$ and $\lambda_1 \geq \sigma^2 \left( \frac{d}{m} - 1\right)$
\end{itemize}
and the value of $\lambda_2$ that achieves zero bias is given by
\begin{align} \label{opt_lambda_2_formula_LS}
    \lambda_2^* = \lambda_1 - \frac{d}{m}\frac{1}{1+\lambda_1/\sigma^2},
\end{align}
where the $S_k$ in $\hat{x}_k = \arg\min_x \|S_kAx-S_kb\|_2^2 + \lambda_2 \|x\|_2^2$ is the Gaussian sketch.
\etheos

\begin{algorithm}[tb]
  \caption{Distributed Randomized Ridge Regression}
  \label{alg:dist_regularized_LS}
\begin{algorithmic}
  \STATE Set $\sigma$ to the mean of singular values of $A$.
  \STATE Calculate $\lambda_2^* = \lambda_1 - \frac{d}{m}\frac{1}{1+\lambda_1/\sigma^2}$.
  \FOR{workers $k=1$ {\bfseries to} $q$ in parallel}
  \STATE Sample $S_k \in \mathbb{R}^{m\times n}$.
  \STATE Compute sketched data $S_kA$ and $S_kb$.
  \STATE Solve $\hat{x}_k = \arg\min_x \|S_kAx-S_kb\|_2^2 + \lambda_2^* \|x\|_2^2$, send to master.
  \ENDFOR
  \STATE \textbf{Master:} return $\bar x = \frac{1}{q} \sum_{k=1}^q \hat x_k$.
\end{algorithmic}
\end{algorithm}

\begin{figure}[ht]
\begin{minipage}[b]{0.48\linewidth}
  \centering
  \centerline{\includegraphics[width=4.4cm]{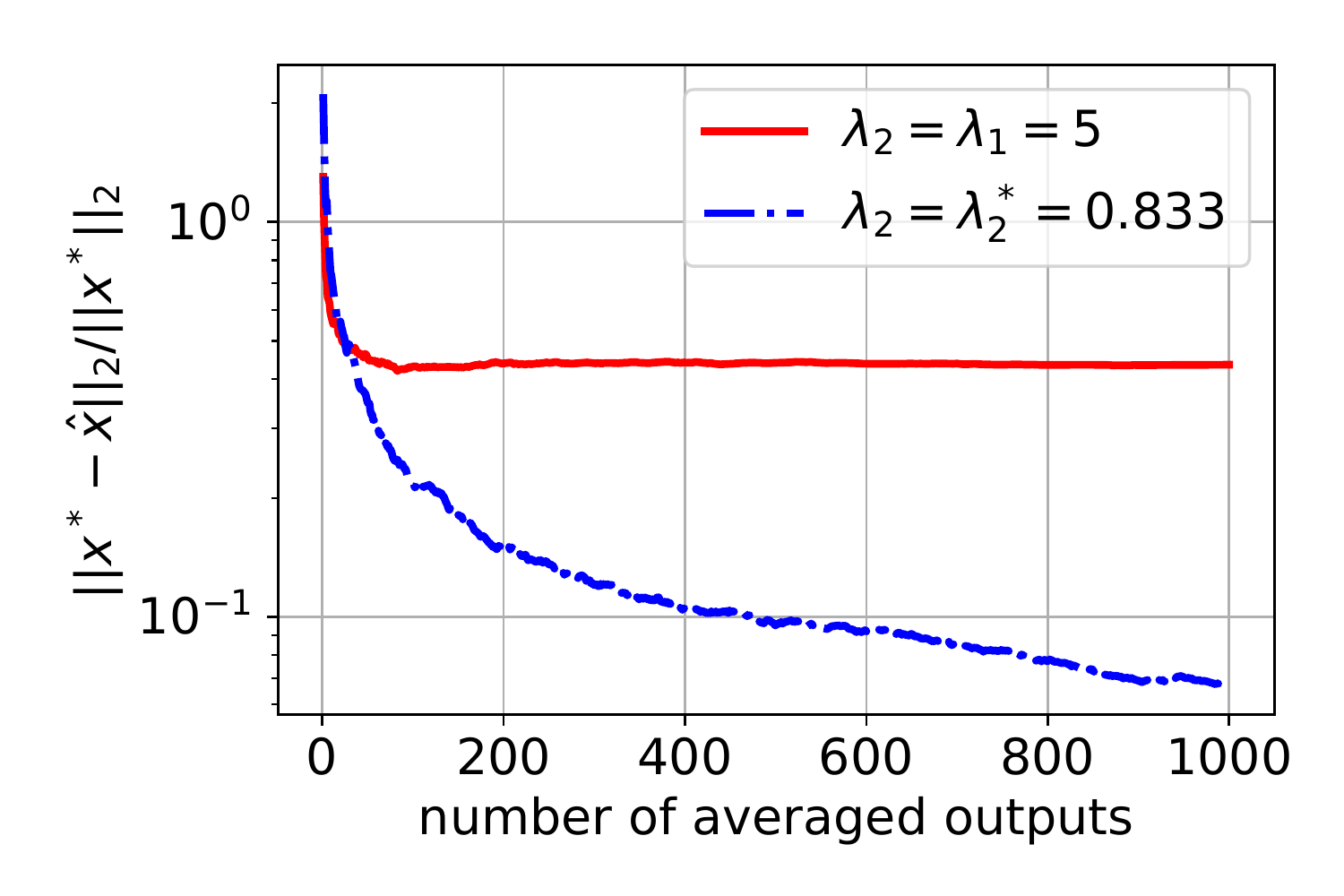}}
  \centerline{(a)}\medskip
\end{minipage}
\hfill
\begin{minipage}[b]{0.48\linewidth}
  \centering
  \centerline{\includegraphics[width=4.4cm]{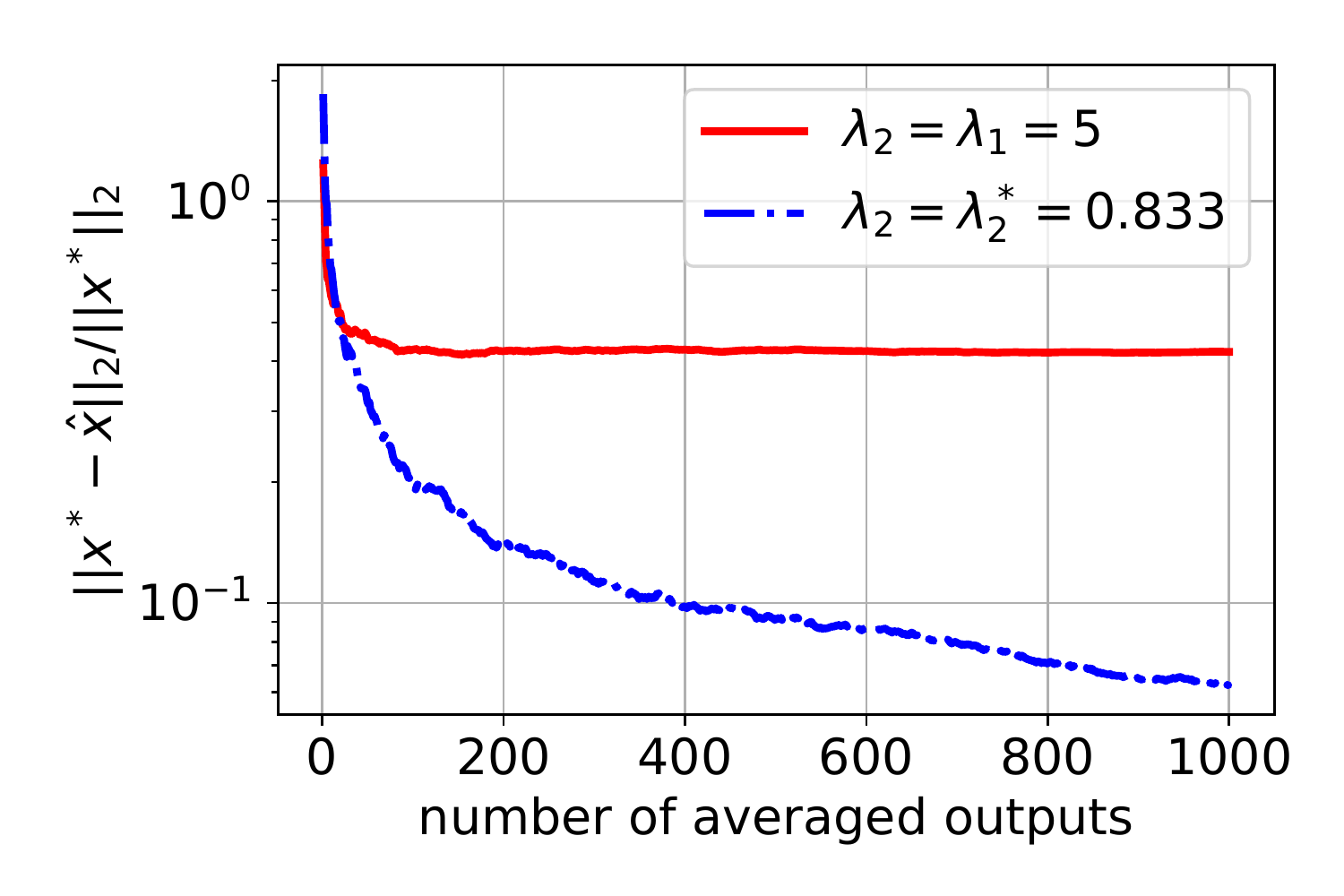}}
  \centerline{(b)}\medskip
\end{minipage}
\caption{Plots of $\|\bar{x}-x^*\|_2 / \|x^*\|_2$ against the number of averaged worker outputs for an unconstrained least squares problem with regularization using Algorithm \ref{alg:dist_regularized_LS}. The dashed blue line corresponds to the case where $\lambda_2$ is determined according to the formula \eqref{opt_lambda_2_formula_LS}, and the solid red line corresponds to the case where $\lambda_2$ is the same as $\lambda_1$. The experimental parameters are as follows: $n=1000$, $d=100$, $\lambda_1=5$, $m=20$, sketch type is Gaussian. (a) All singular values of $A$ are $1$, (b) Singular values of $A$ are not identical and their mean is $1$.}
\label{fig:reg_ls_gaussketch}
\end{figure}

Figure \ref{fig:reg_ls_gaussketch} illustrates the implications of Theorem \ref{opt_lambda_2_newton_LS}. If $\lambda_2$ is chosen according to the formula in \eqref{opt_lambda_2_formula_LS}, then the averaged solution $\bar{x}$ is a better approximation to $x^*$ than if we had used $\lambda_2=\lambda_1$. The data matrix $A$ in Figure \ref{fig:reg_ls_gaussketch}(a) has identical singular values, and \ref{fig:reg_ls_gaussketch}(b) shows the case where the singular values of $A$ are not identical. When the singular values of $A$ are not all equal to each other, we set $\sigma$ to the mean of the singular values of $A$ as a heuristic, which works extremely well as shown in the figure.
According to the formula \eqref{opt_lambda_2_formula_LS}, the value of $\lambda_2$ that we need to use to achieve zero bias is found to be $\lambda_2^*=0.833$ whereas $\lambda_1 = 5$. The plot in Figure \ref{fig:reg_ls_gaussketch}(b) illustrates that even if the assumption that $\Sigma=\sigma I_d$ in Theorem \ref{opt_lambda_2_newton_LS} is violated, the proposed bias corrected averaging method outperforms vanilla averaging \cite{mahoney2018averaging} where $\lambda_2=\lambda_1$.

\subsection{Varying Sketch Sizes}
Let us now consider the scenario where we have different sketch sizes in each worker. This situation frequently arises in heterogeneous computing environments. Specifically, let us assume that the sketch size for worker $k$ is $m_k$, $k=1,2,...,q$. By Theorem \ref{opt_lambda_2_newton_LS}, by choosing the regularization parameter for worker $k$ as
\begin{align*}
    \lambda_2^*(k) = \lambda_1 - \frac{d}{m_k}\frac{1}{1+\lambda_1/\sigma^2},
\end{align*}
it is possible to obtain unbiased estimators $\hat{x}_k$ for $k=1,...,q$ and hence an unbiased averaged result $\bar{x}$. Note that here we assume that the sketch size for each worker satisfies the condition in Theorem \ref{opt_lambda_2_newton_LS} for zero bias in each estimator $\hat{x}_k$, that is, either $m_k > d$ or $m_k \leq d$ and $\lambda_1 \geq \sigma^2 (d/m_k-1)$.

\section{Distributed Newton Sketch} \label{sec:newton_sketch}
We have considered the linear least squares problem without and with regularization in Sections \ref{sec:ihs} and \ref{sec:newton_sketch}, respectively. Next, we consider randomized second order methods for solving a broader range of problems, where we consider the distributed version of the Newton Sketch algorithm described in \cite{pilanci2017newton}. We consider Hessian matrices of the form $H_t=(H_t^{1/2})^T H_t^{1/2}$, where we assume that $H_t^{1/2} \in \mathbb{R}^{n\times d}$ is a full rank matrix and $n \geq d$. Note that this factorization is already available in terms of scaled data matrices in many problems as we illustrate in the sequel. This enables the fast construction of an approximation of $H_t$ by applying sketching $S_tH_t^{1/2}$ which leads to the approximation $\hat{H}_t = (S_tH_t^{1/2})^T S_tH_t^{1/2}$.
Averaging in the case of Hessian matrices of the form $H_t=(H_t^{1/2})^T H_t^{1/2} + \lambda_1 I_d$ (i.e. regularized) will be considered in the next section.

Let us consider the updates in classical Newton's method:
\begin{align}
    x_{t+1} = x_t - \alpha_1 H_t^{-1} g_t,
\end{align}
where $H_t \in \mathbb{R}^{d\times d}$ and $g_t \in \mathbb{R}^{d}$ denote the Hessian matrix and the gradient vector at iteration $t$ respectively, and $\alpha_1$ is the step size. In contrast, Newton Sketch performs the approximate updates
\begin{align}
    x_{t+1} = x_t + \alpha_1 \arg\min_{\Delta}(\frac{1}{2}\|S_t H_t^{1/2}\Delta \|_2^2 + g_t^T\Delta),
\end{align}
where the sketching matrices $S_t \in \mathbb{R}^{m\times n}$ are refreshed every iteration. There is a multitude of options for distributing Newton's method and Newton Sketch. Here we consider a scheme that is similar in spirit to the GIANT algorithm \cite{mahoney2018giant} where workers communicate approximate length-$d$ update directions to be averaged at the master node. Another alternative scheme would be to communicate the approximate Hessian matrices, which would require an increased communication load of $d^2$ numbers.

The updates for distributed Newton sketch are given by
\begin{align}
    x_{t+1} = x_t + \alpha_2 \frac{1}{q} \sum_{k=1}^q \arg\min_{\Delta}\,\frac{1}{2}\|S_{t,k} H_t^{1/2}\Delta \|_2^2 + g_t^T\Delta.
\end{align}
Note that the above update requires access to the full gradient $g_t$. If workers do not have access to the entire dataset, then this requires an additional communication round per iteration where workers communicate their local gradients with the master node, which computes the full gradient and broadcasts to workers. The details of the distributed Newton Sketch method is given in Algorithm \ref{alg:dist_newton_sketch}. 

\begin{algorithm}[tb]
  \caption{Distributed Newton Sketch}
  \label{alg:dist_newton_sketch}
\begin{algorithmic}
  \STATE {\bfseries Input:} Tolerance $\epsilon$
  \REPEAT
  \FOR{workers $k=1$ {\bfseries to} $q$ (in parallel)}
  \STATE Sample $S_{t,k} \in \mathbb{R}^{m\times n}$.
  \STATE Sketch $S_{t,k}H_t^{1/2}$.
  \STATE Obtain the gradient $g_t$.
  \STATE Compute approximate Newton direction $\hat{\Delta}_{k,t} = \arg\min_{\Delta}(\frac{1}{2}\|S_{t,k} H_t^{1/2}\Delta \|_2^2 + g_t^T\Delta)$ and send to master.
  \ENDFOR
  \STATE \textbf{Master:} Determine $\alpha_2$ and update $x_{t+1} = x_t + \alpha_2 \frac{1}{q} \sum_{k=1}^q \hat{\Delta}_{k,t}$.
  \UNTIL{$g_t^T \left(\sum_{k=1}^q \hat{\Delta}_{k,t}\right)/2 \geq \epsilon$ is satisfied}
\end{algorithmic}
\end{algorithm}

\subsection{Gaussian Sketch}
We analyze the bias and the variance of the update directions for distributed Newton sketch, and give exact expressions for Gaussian sketching matrices. 

Let $\Delta_t^*$ denote the exact Newton update direction at iteration $t$, then
\begin{align}
    \Delta_t^* = ((H_t^{1/2})^T H_t^{1/2})^{-1}g_t
\end{align}
and let $\hat{\Delta}_{k,t}$ denote the approximate update direction outputted by worker $k$ at iteration $t$, which is given by
\begin{align}
    \hat{\Delta}_{k,t} = \alpha_s ((H_t^{1/2})^TS_{t,k}^TS_{t,k}H_t^{1/2})^{-1}g_t.
\end{align}
Note that the step size for the averaged update direction will be calculated as $\alpha_2 = \alpha_1 \alpha_s$. Theorem \ref{thm:dist_newton_sketch} characterizes how the update directions needs to be modified to obtain an unbiased update direction, and a minimum variance estimator for the update direction.

\btheos[] \label{thm:dist_newton_sketch}
For Gaussian sketches $S_{t,k}$, assuming $H_t^{1/2}$ is full column rank, the variance $\Exs[\|H_t^{1/2}(\hat{\Delta}_{k,t} - \Delta_t^*)\|_2^2]$ is minimized when $\alpha_s$ is chosen as $\alpha_s =  \frac{\theta_1}{\theta_2}$ whereas the bias $\Exs[H_t^{1/2}(\hat{\Delta}_{k,t} - \Delta_t^*)]$ is zero when $\alpha_s =  \frac{1}{\theta_1}$, where $\theta_1$ and $\theta_2$ are as defined in \eqref{eq:theta_1_2_definitions}.
\etheos

\begin{figure}[ht]
\begin{minipage}[b]{0.48\linewidth}
  \centering
  \centerline{\includegraphics[width=4.3cm]{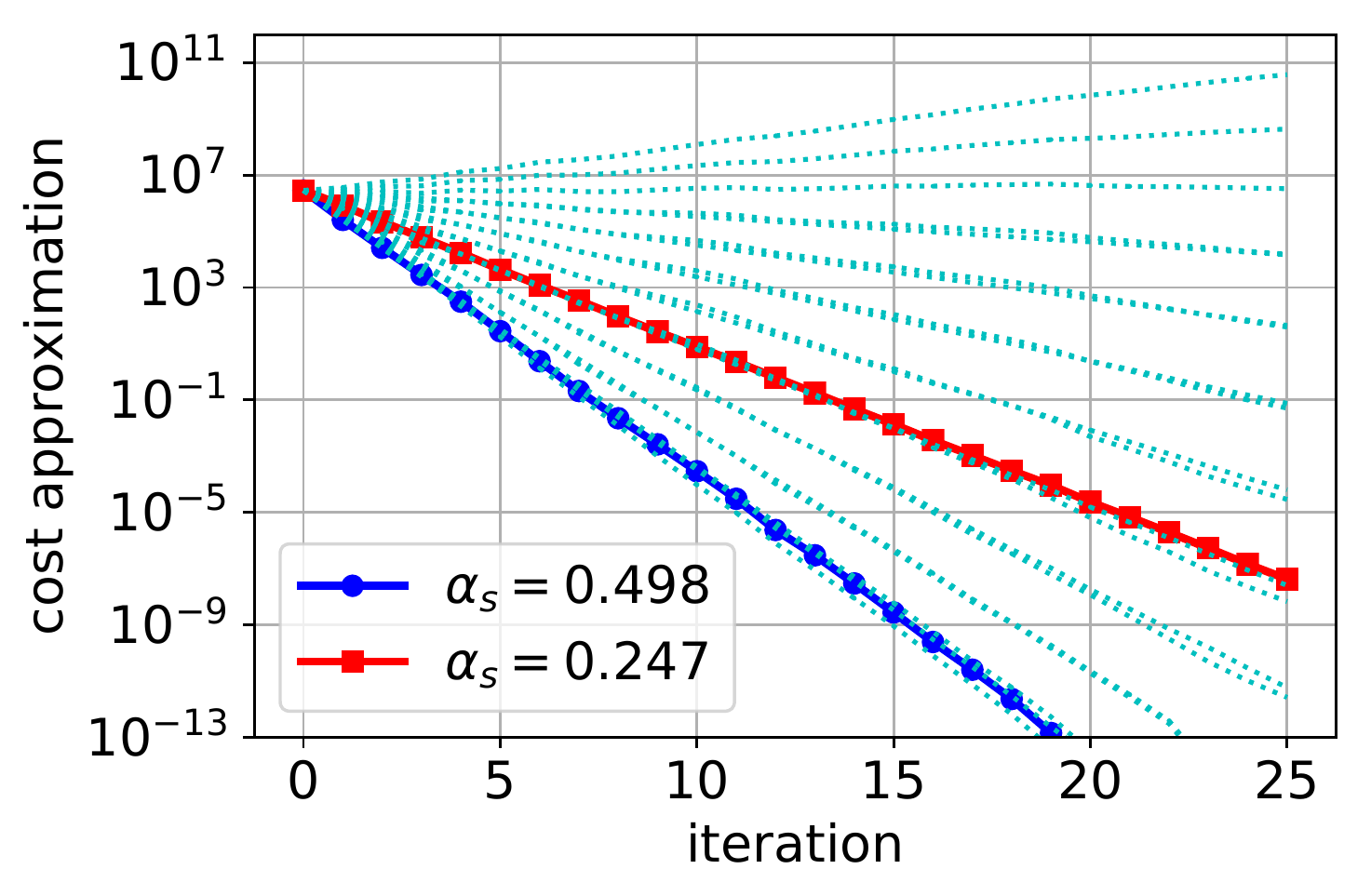}}
  \centerline{(a)}\medskip
\end{minipage}
\hfill
\begin{minipage}[b]{0.48\linewidth}
  \centering
  \centerline{\includegraphics[width=4.3cm]{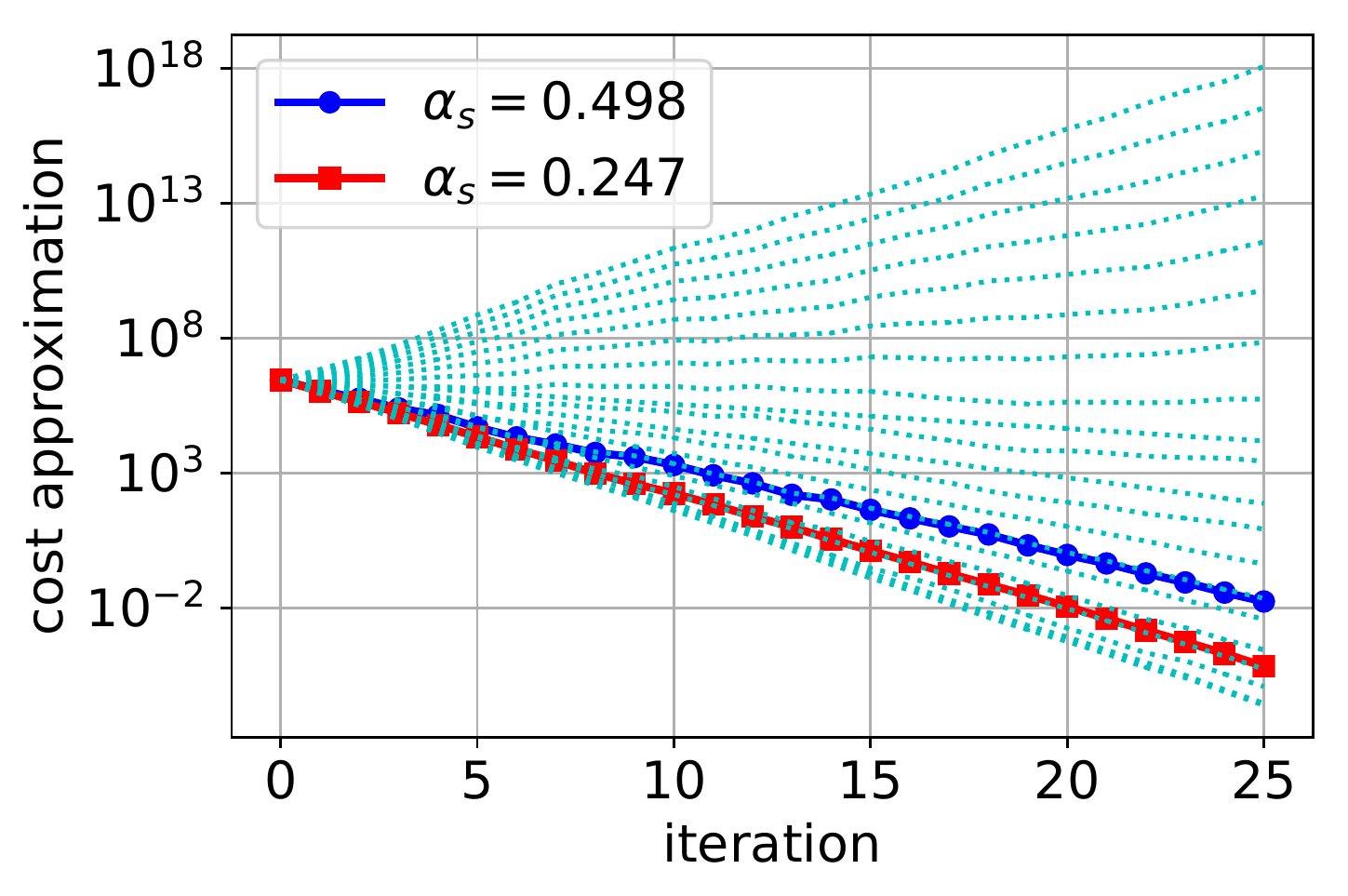}}
  \centerline{(b)}\medskip
\end{minipage}
\caption{Cost approximation $(f(x_t)-f(x^*))/f(x^*)$ for Algorithm \ref{alg:dist_newton_sketch} against iteration number $t$ for various step sizes in solving a linear least squares problem on randomly generated data. The cyan colored dotted lines show cost approximation when we make a search for the learning rate $\alpha_s$ between $0.05$ and $1$. The blue line with circles corresponds to $\alpha_s$ that leads to the unbiased estimator and the red line with squares corresponds to $\alpha_s$ that gives the minimum variance. The step size scaling factors $\alpha_s$ were calculated using the formulas in Theorem \ref{thm:dist_newton_sketch} and are marked on the plots. The parameters used in these experiments are $n=1000$, $d=200$, $m=400$. (a) $q=10$ workers, (b) $q=2$ workers.}
\label{fig:newton_sketch_data_gauss4th}
\end{figure}

Figure \ref{fig:newton_sketch_data_gauss4th} demonstrates that choosing $\alpha_2=\alpha_1 \alpha_s$ where $\alpha_s$ is calculated using the unbiased estimator formula leads to faster decrease of the objective value when the number of workers is large. If the number of workers is small, one should choose the step size that minimizes variance instead. Figure \ref{fig:newton_sketch_data_gauss4th}(a) illustrates that the blue curve with squares is in fact the best one could hope to achieve as it is very close to the best cyan dotted line.

\textit{Non-identical sketch sizes:} Theorem \ref{thm:dist_newton_sketch} establishes that whenever the sketch dimension varies among workers, it is possible to obtain an unbiased update direction by computing $\alpha_s$ for every worker individually.

\section{Distributed Newton Sketch for Regularized Problems} \label{sec:regularized_newton}
We now consider problems with $\ell_2$ regularization. In particular, we study problems whose Hessian matrices are of the form $H_t=(H_t^{1/2})^TH_t^{1/2} + \lambda_1 I_d$. Sketching can be applied to obtain approximate Hessian matrices as $H_t=(S_tH_t^{1/2})^TS_tH_t^{1/2} + \lambda_2 I_d$. Note that the case $\lambda_2=\lambda_1$ corresponds to the setting in the GIANT algorithm described in \cite{mahoney2018giant}. 

Theorem \ref{opt_lambda_2_newton} establishes that $\lambda_2$ should be chosen according to the formula \eqref{opt_lambda_2_formula} under the assumption that the singular values of $H_t^{1/2}$ are identical. We later verify empirically that when the singular values are not identical, plugging the mean of the singular values into the formula still leads to improvements over the case of $\lambda_2 = \lambda_1$.

\btheos[] \label{opt_lambda_2_newton}
Given the thin SVD decomposition $H_t^{1/2}=U\Sigma V^T \in \mathbb{R}^{n\times d}$ and $n \geq d$ where $H_t^{1/2}$ is assumed to have full rank and satisfy $\Sigma = \sigma I_d$, the bias of the single-sketch Newton step estimator $\Exs[ H_t^{1/2}(\hat{\Delta}_{t,k} - \Delta^*)]$ is equal to zero as $n$ goes to infinity when $\lambda_2$ is chosen as
\begin{align} \label{opt_lambda_2_formula}
    \lambda_2^* = \frac{\lambda_1 + \sigma^2\frac{d}{m}}{1 + \frac{d}{m} \frac{1}{1 + (\lambda_1 / \sigma^2)}},
\end{align}
where $\Delta^* = ((H_t^{1/2})^TH_t + \lambda_1 I_d)^{-1} g$ and $\hat{\Delta}_k = ((S_{t,k}H_t^{1/2})^TS_{t,k}H_t + \lambda_2 I_d)^{-1} g$, and $S_{t,k}$ is the Gaussian sketch.
\etheos

\section{Example Applications} \label{sec:example_problems}
In this section, we describe examples where our methodology can be applied. In particular, the problems in this section are convex problems that are efficiently addressed by our methods in distributed systems. We present numerical results on these problems in the next section.

\subsection{Logistic Regression} 
Let us consider the logistic regression problem with $\ell_2$ penalty given by $\textrm{minimize}_{x} f(x)$ where
\begin{align}
     f(x) = -\sum_{i=1}^n \left( y_i \log(p_i) + (1-y_i)\log(1-p_i)  \right) + \frac{\lambda_1}{2} \|x\|_2^2,
\end{align}
and $p \in \mathbb{R}^{n}$ is defined such that $p_i = 1/(1+\exp(-\tilde{a}_i^Tx))$. $\tilde{a}_i$ represents the $i$'th row of the data matrix $A \in \mathbb{R}^{n\times d}$. The output vector is denoted by $y \in \mathbb{R}^n$.

The gradient and Hessian for $f(x)$ are as follows
\begin{align*}
    g = A^T(p-y) + \lambda_1 x, \\
    H = A^TDA + \lambda_1I_d,
\end{align*}
$D$ is a diagonal matrix with the entries of the vector $p(1-p)$ as its diagonal entries. The sketched Hessian matrix in this case can be formed as
$(SD^{1/2}A)^T(SD^{1/2}A) + \lambda_2^* I_d$ and $\lambda_2^*$ can be calculated using \eqref{opt_lambda_2_formula}, setting $\sigma$ the mean of singular values of $D^{1/2}A$ in the formula. Because $D$ changes every iteration, it might be computationally infeasible to re-compute the mean of the singular values. However, we have found through experiments that it is not required to compute the exact value of the mean of the singular values. For instance, setting $\sigma$ to the mean of the diagonals of $D^{1/2}$ as a heuristic works sufficiently well.

\subsection{Inequality Constrained Optimization} 
Next, we consider the following inequality constrained optimization problem,
\begin{align}
    \textrm{minimize}_{x} \quad & ||x-c||_2^2 \nonumber \\
    \textrm{subject to} \quad & ||Ax||_{\infty} \leq \lambda 
\end{align}
where $A \in \mathbb{R}^{n \times d}$, and $c \in \mathbb{R}^{d}$ are the problem data, and $\lambda \in \mathbb{R}$ is a positive scalar. Note that this problem is the dual of the Lasso problem given by $\text{min}_x$ $\lambda \|x\|_1 + \frac{1}{2}\|Ax-c\|_2^2$.

The above problem can be tackled by the standard log-barrier method \cite{boyd2004convex}, by solving sequences of unconstrained barrier penalized problems as follows
\begin{align} \label{log_barrier_opt_prob_2}
    \textrm{minimize}_x \, - \sum_{i=1}^n \log(-\tilde{a}_i^Tx+\lambda) -\sum_{i=1}^n \log(\tilde{a}_i^Tx+\lambda) \nonumber \\ + \lambda_1 ||x||_2^2 - 2\lambda_1 c^Tx + \lambda_1 ||c||_2^2
\end{align}
where $\tilde{a}_i$ represents the $i$'th row of $A$.
The gradient and Hessian of the objective are given by
\begin{align*}
    g &= -A_c^T D \mathbf{1}_{2n\times 1} + 2\lambda_1 x - 2\lambda_1 c, \\
    H &= (DA_c)^T(DA_c) + 2\lambda_1 I_d.
\end{align*}
Here $A_c = [A^T, -A^T]^T$ and $D$ is a diagonal matrix with the element-wise inverse the vector $(A_cx-\mathbf{1}_{2n\times 1})$ as its diagonal entries. $\mathbf{1}_{2n\times 1}$ is a length-$2n$ vector of all $1$'s. The sketched Hessian can be written in the form of $(SDA_c)^T(SDA_c) + \lambda_2I_d$.

\textit{Remark:} Since we have the term $2\lambda_1 I_d$ in $H$ (instead of $\lambda_1 I_d$), we need to plug in $2\lambda_1$ instead of $\lambda_1$ in the formula for computing $\lambda_2^*$. 

\section{Numerical Results} \label{sec:numerical_results}

\subsection{Distributed Iterative Hessian Sketch}
We have evaluated the distributed IHS algorithm on the serverless computing platform AWS Lambda. 
In the implementation, each serverless function is responsible for solving one sketched problem per iteration. Workers wait once they finish their computation for that iteration until the next iterate $x_{t+1}$ becomes available. The master node, which is another AWS Lambda worker, is responsible for collecting and averaging the worker outputs and broadcasting the next iterate $x_{t+1}$.

\begin{figure}[ht]
\begin{center}
\centerline{\includegraphics[width=\columnwidth]{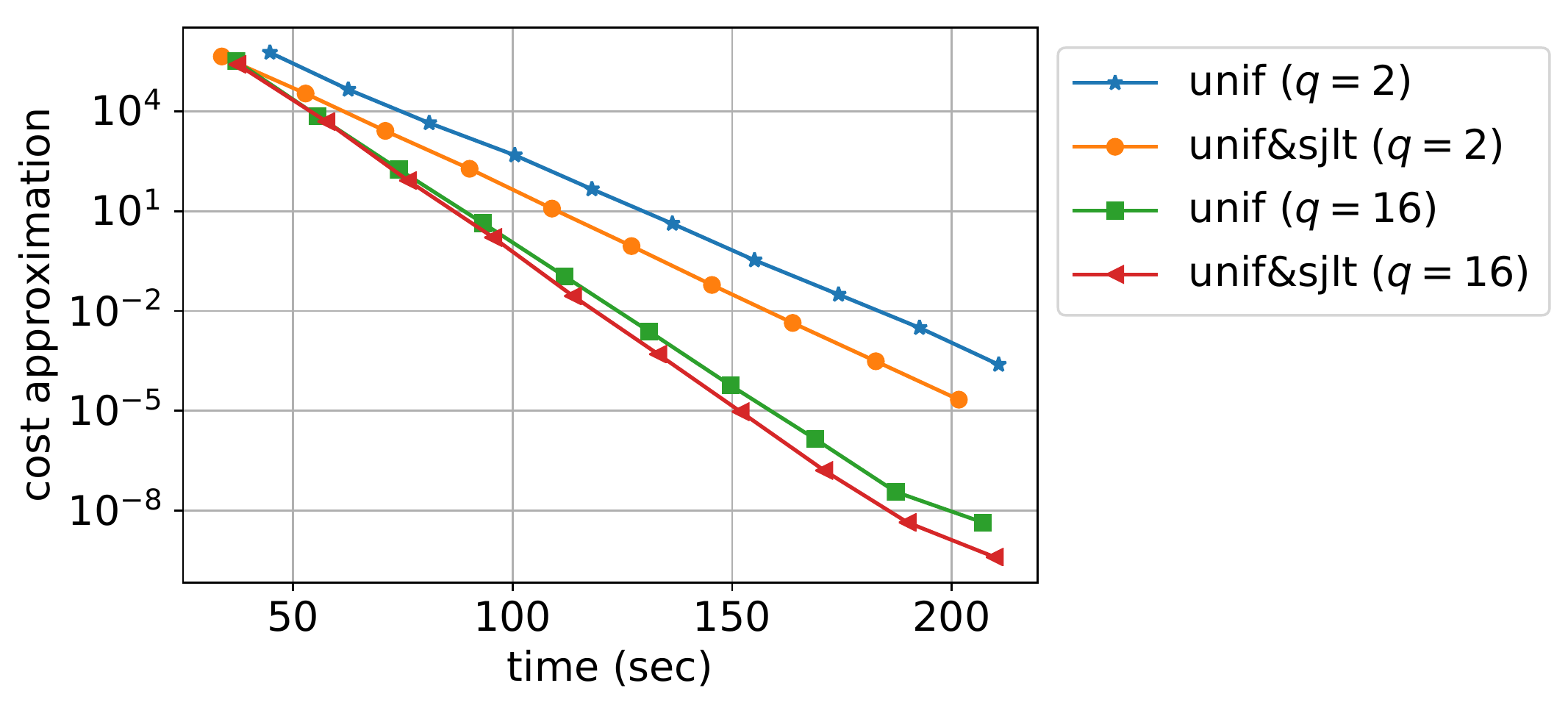}}
\caption{Cost approximation $(f(x_t)-f(x^*))/f(x^*)$ vs time for the distributed IHS algorithm running on AWS Lambda for solving the linear least squares problem given in \eqref{eq:standard_least_sq} for randomly generated data. Unif is short for uniform sampling and unif\&sjlt is short for hybrid sketch where uniform sampling is followed by SJLT. Problem parameters are as follows: $n=250000$, $d=500$, $m=6000$, $m_2=20000$, and $q$ as specified in the legend.}
\label{fig:IHS_plot_aws_lambda}
\end{center}
\end{figure}

Figure \ref{fig:IHS_plot_aws_lambda} shows the scaled difference between the cost for the $t$'th iterate and the optimal cost (i.e. $(f(x_t)-f(x^*))/f(x^*)$) versus iteration number $t$ for the distributed IHS algorithm given in Algorithm \ref{alg:dist_ihs}. Due to the relatively small size of the problem, we have each worker compute the exact gradient without requiring an additional communication round per iteration to form the full gradient. We note that, in problems where it's not feasible for workers to form the full gradient due to limited memory, one can include an additional communication round where each worker sends their local gradient to the master node, and the master node forms the full gradient and distributes it to the worker nodes. 

\subsection{Inequality Constrained Optimization}
\begin{figure}[ht]
\begin{center}
\centerline{\includegraphics[width=0.9\columnwidth]{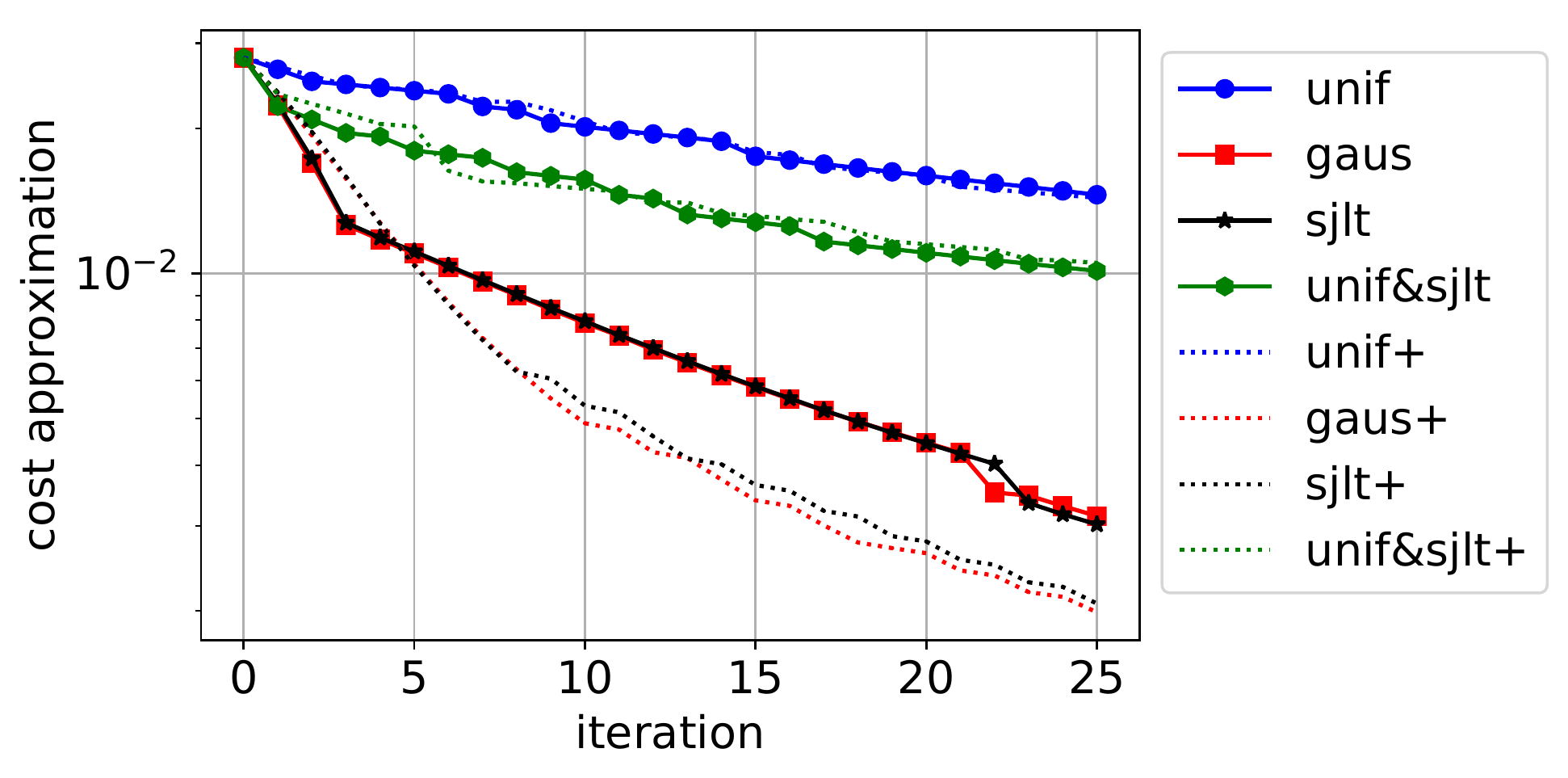}}
\caption{Plot shows cost approximation of the iterate $x_t$ (i.e., $(f(x_t)-f(x^*))/f(x^*)$) against iteration number $t$ for various sketches in solving an inequality constrained optimization problem, namely, the problem given in \eqref{log_barrier_opt_prob_2}. Abbreviations used in the plot are as follows. Unif: Uniform sampling, gaus: Gaussian sketch, unif\&sjlt: Hybrid sketch where uniform sampling is followed by SJLT. The abbreviations followed by $+$'s refer to the bias corrected versions. Problem parameters are as follows: $n=500$, $d=200$, $\lambda_1=1000$, $m=50$, $m_2=8m=400$, $q=10$, $\lambda=0.01$, $s=10$. }
\label{fig:log_barrier_experiment}
\end{center}
\end{figure}

Figure \ref{fig:log_barrier_experiment} compares various sketches with and without bias correction for the distributed Newton sketch algorithm when it is used to solve the problem given in \eqref{log_barrier_opt_prob_2}. For each sketch, we have plotted the performance for $\lambda_2=\lambda_1$ and the bias corrected version $\lambda_2=\lambda_2^*$. The bias corrected versions are shown as the dotted lines.
In these experiments, we have set $\sigma$ to the minimum of the singular values of $DA$ as we have observed that setting $\sigma$ to the minimum of the singular values of $DA$ performed better than setting it to their mean. 

Even though we have derived the bias correction formula for Gaussian sketch, we observe that it improves the performance of SJLT as well. We see that Gaussian sketch and SJLT perform the best out of the 4 sketches we have experimented with. We note that computational complexity of sketching for SJLT is lower than it is for Gaussian sketch, and hence the natural choice would be to use SJLT in this case. 

\subsection{Scalability of the Serverless Implementation}
\begin{figure}[ht]
\begin{center}
\centerline{\includegraphics[width=0.75\columnwidth]{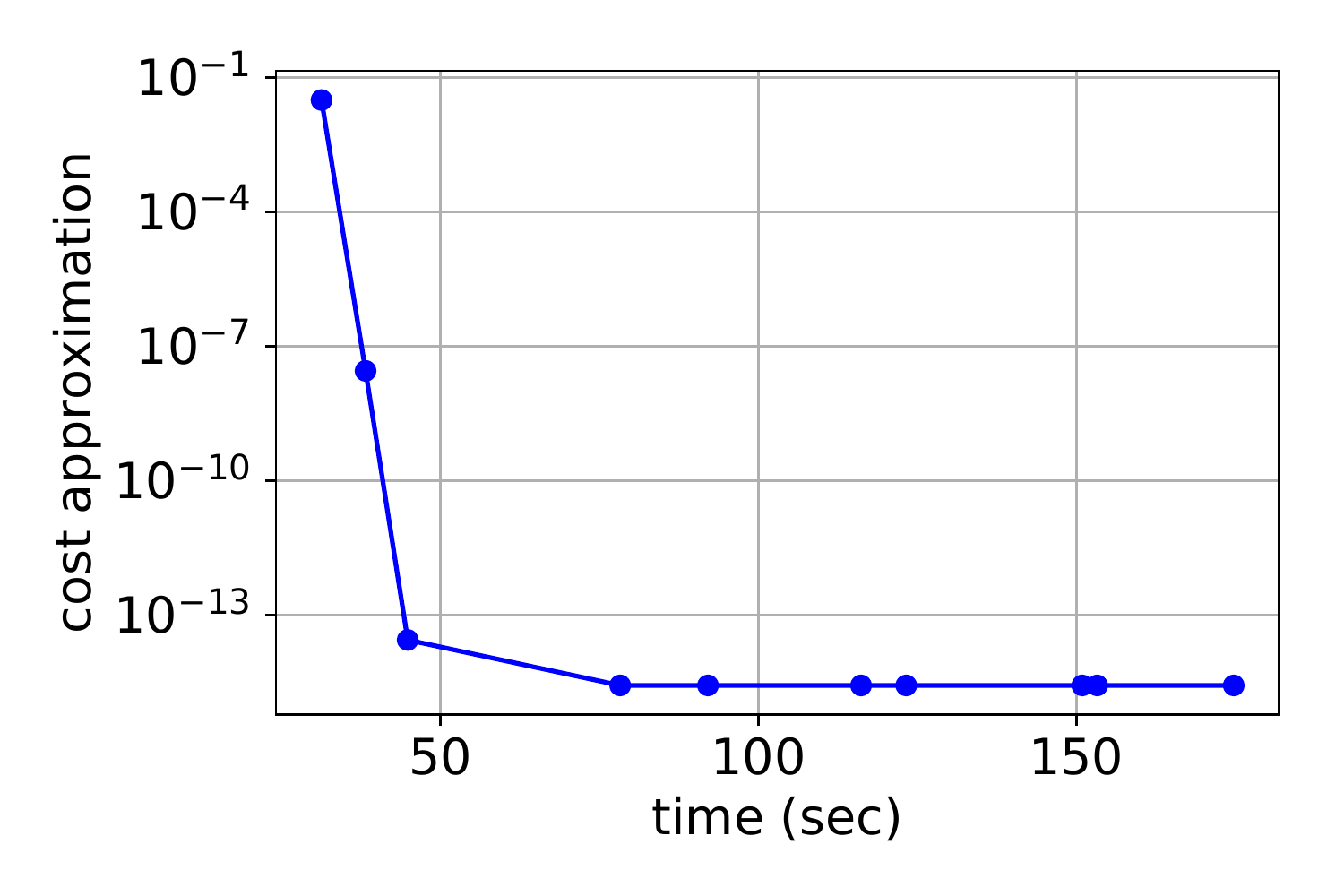}}
\caption{Cost approximation vs time when we solve the problem given in \eqref{log_barrier_opt_prob_2}) for a large scale randomly generated dataset ($44$ GB sized) on AWS Lambda. Circles correspond to times that iterates $x_t$ are computed.  Problem parameters are as follows: $n=200000$, $d=30000$, $\lambda_1=1$, $m=2000$, $q=100$, $\lambda=10$.}
\label{fig:log_barrier_awslambda_scalability}
\end{center}
\end{figure}

Figure \ref{fig:log_barrier_awslambda_scalability} shows the cost against time when we solve the problem given in \eqref{log_barrier_opt_prob_2} for large scale data on AWS Lambda using the distributed Newton sketch algorithm. The setting in this experiment is such that each worker has access to a different subset of data, and there is no additional sketching applied. The dataset used is randomly generated and the goal here is to demonstrate the scalability of the algorithm and the serverless implementation. The size of the data matrix $A$ is $44$ GB.

In the serverless implementation, we reuse the serverless functions during the course of the algorithm, meaning that the same $q=100$ functions are used for every iteration. We note that every iteration requires two rounds of communication with the master node. The first round is for the communication of the local gradients, and the second round is for the approximate update directions. The master node, also a serverless function, is also reused across iterations. Figure \ref{fig:log_barrier_awslambda_scalability} illustrates that each iteration takes a different amount of time and iteration times can be as short as $5$ seconds. The reason for some iterations taking much longer times is what is referred to as the straggler problem, which is a phenomenon commonly encountered in distributed computing. More precisely, the iteration time is determined by the slowest of the $q=100$ nodes and nodes often slow down for a variety of reasons causing stragglers. A possible solution to the issue of straggling nodes is to use error correcting codes to insert redundancy to computation and hence to avoid waiting for the outputs of all of the worker nodes \cite{Lee2018}. We identify that implementing straggler mitigation for solving large scale problems via approximate second order optimization methods such as distributed Newton sketch is a promising direction.

\section{Conclusion} \label{sec:conclusion}
In this work, we have studied averaging for a wide class of randomized second order algorithms. Sections \ref{sec:ihs} and \ref{sec:regularized_least_squares} are focused on the problem of linear least squares whereas the results of sections \ref{sec:newton_sketch} and \ref{sec:regularized_newton} are applicable to a more general class of problems. We have shown that for problems involving regularization, averaging requires more detailed analysis compared to problems without regularization. When the regularization term is not scaled properly, the resulting estimators are biased, and averaging a large number of independent sketched solutions does not converge to the true solution. We have provided closed-form formulas for scaling the regularization coefficient to obtain unbiased estimators. This method does not demand any additional computational cost, while guaranteeing convergence to the optimum. We also extended our analysis to non-identical sketch dimensions for heterogeneous computing environments.

A future research direction is the analysis of the bias and variance, and unbiased schemes for a broader class of sketching matrices under less restrictive assumptions on the singular values as in Theorems \ref{opt_lambda_2_newton_LS} and  \ref{opt_lambda_2_newton}.

\clearpage
\bibliography{REFS_1,REFS_2}

\begin{thebibliography}{19}
\providecommand{\natexlab}[1]{#1}
\providecommand{\url}[1]{\texttt{#1}}
\expandafter\ifx\csname urlstyle\endcsname\relax
  \providecommand{\doi}[1]{doi: #1}\else
  \providecommand{\doi}{doi: \begingroup \urlstyle{rm}\Url}\fi

\bibitem[Avron et~al.(2010)Avron, Maymounkov, and Toledo]{avron2010blendenpik}
Avron, H., Maymounkov, P., and Toledo, S.
\newblock Blendenpik: Supercharging lapack's least-squares solver.
\newblock \emph{SIAM Journal on Scientific Computing}, 32\penalty0
  (3):\penalty0 1217--1236, 2010.

\bibitem[Boyd \& Vandenberghe(2004)Boyd and Vandenberghe]{boyd2004convex}
Boyd, S. and Vandenberghe, L.
\newblock \emph{Convex optimization}.
\newblock Cambridge university press, 2004.

\bibitem[Drineas \& Mahoney(2016)Drineas and Mahoney]{drineas2016randnla}
Drineas, P. and Mahoney, M.~W.
\newblock Rand{NLA}: randomized numerical linear algebra.
\newblock \emph{Communications of the ACM}, 59\penalty0 (6):\penalty0 80--90,
  2016.

\bibitem[Drineas et~al.(2011)Drineas, Mahoney, Muthukrishnan, and
  Sarl{\'o}s]{drineas2011faster}
Drineas, P., Mahoney, M.~W., Muthukrishnan, S., and Sarl{\'o}s, T.
\newblock Faster least squares approximation.
\newblock \emph{Numerische mathematik}, 117\penalty0 (2):\penalty0 219--249,
  2011.

\bibitem[Lacotte \& Pilanci(2019)Lacotte and Pilanci]{lacotte2019faster}
Lacotte, J. and Pilanci, M.
\newblock Faster least squares optimization, 2019.

\bibitem[Lee et~al.(2018)Lee, Lam, Pedarsani, Papailiopoulos, and
  Ramchandran]{Lee2018}
Lee, K., Lam, M., Pedarsani, R., Papailiopoulos, D., and Ramchandran, K.
\newblock Speeding up distributed machine learning using codes.
\newblock \emph{IEEE Transactions on Information Theory}, 64\penalty0
  (3):\penalty0 1514--1529, March 2018.

\bibitem[Liu \& Dobriban(2019)Liu and Dobriban]{liu2019ridge}
Liu, S. and Dobriban, E.
\newblock Ridge regression: Structure, cross-validation, and sketching, 2019.

\bibitem[Mahoney(2011)]{mahoney2011randomized}
Mahoney, M.~W.
\newblock Randomized algorithms for matrices and data.
\newblock \emph{Foundations and Trends{\textregistered} in Machine Learning},
  3\penalty0 (2):\penalty0 123--224, 2011.

\bibitem[Nelson \& Nguy{\^e}n(2013)Nelson and Nguy{\^e}n]{nelson2013osnap}
Nelson, J. and Nguy{\^e}n, H.~L.
\newblock Osnap: Faster numerical linear algebra algorithms via sparser
  subspace embeddings.
\newblock In \emph{Foundations of Computer Science (FOCS), 2013 IEEE 54th
  Annual Symposium on}, pp.\  117--126. IEEE, 2013.

\bibitem[Pilanci \& Wainwright(2015)Pilanci and
  Wainwright]{pilanci2015randomized}
Pilanci, M. and Wainwright, M.~J.
\newblock Randomized sketches of convex programs with sharp guarantees.
\newblock \emph{IEEE Transactions on Information Theory}, 61\penalty0
  (9):\penalty0 5096--5115, 2015.

\bibitem[Pilanci \& Wainwright(2016)Pilanci and Wainwright]{PilWai14b}
Pilanci, M. and Wainwright, M.~J.
\newblock Iterative hessian sketch: Fast and accurate solution approximation
  for constrained least-squares.
\newblock \emph{The Journal of Machine Learning Research}, 17\penalty0
  (1):\penalty0 1842--1879, 2016.

\bibitem[Pilanci \& Wainwright(2017)Pilanci and Wainwright]{pilanci2017newton}
Pilanci, M. and Wainwright, M.~J.
\newblock Newton sketch: A near linear-time optimization algorithm with
  linear-quadratic convergence.
\newblock \emph{SIAM Journal on Optimization}, 27\penalty0 (1):\penalty0
  205--245, 2017.

\bibitem[{Reisizadeh} et~al.(2017){Reisizadeh}, {Prakash}, {Pedarsani}, and
  {Avestimehr}]{avestimehr2017heterogenous_coded}
{Reisizadeh}, A., {Prakash}, S., {Pedarsani}, R., and {Avestimehr}, S.
\newblock Coded computation over heterogeneous clusters.
\newblock In \emph{2017 IEEE International Symposium on Information Theory
  (ISIT)}, pp.\  2408--2412, June 2017.
\newblock \doi{10.1109/ISIT.2017.8006961}.

\bibitem[Rokhlin et~al.(2009)Rokhlin, Szlam, and Tygert]{rokhlin2009randomized}
Rokhlin, V., Szlam, A., and Tygert, M.
\newblock A randomized algorithm for principal component analysis.
\newblock \emph{SIAM Journal on Matrix Analysis and Applications}, 31\penalty0
  (3):\penalty0 1100--1124, 2009.

\bibitem[Vempala(2005)]{vempala2005random}
Vempala, S.~S.
\newblock \emph{The random projection method}, volume~65.
\newblock American Mathematical Soc., 2005.

\bibitem[Wang et~al.(2018{\natexlab{a}})Wang, Tan, Chen, Lin, Keerthi, Mahajan,
  Sundararajan, and Lin]{wang2018distributed}
Wang, C.-C., Tan, K.~L., Chen, C.-T., Lin, Y.-H., Keerthi, S.~S., Mahajan, D.,
  Sundararajan, S., and Lin, C.-J.
\newblock Distributed newton methods for deep neural networks.
\newblock \emph{Neural computation}, \penalty0 (Early Access):\penalty0 1--52,
  2018{\natexlab{a}}.

\bibitem[Wang et~al.(2017)Wang, Gittens, and Mahoney]{mahoney2018averaging}
Wang, S., Gittens, A., and Mahoney, M.~W.
\newblock Sketched ridge regression: Optimization perspective, statistical
  perspective, and model averaging.
\newblock \emph{J. Mach. Learn. Res.}, 18\penalty0 (1):\penalty0 8039--8088,
  January 2017.

\bibitem[Wang et~al.(2018{\natexlab{b}})Wang, Roosta, Xu, and
  Mahoney]{mahoney2018giant}
Wang, S., Roosta, F., Xu, P., and Mahoney, M.~W.
\newblock Giant: Globally improved approximate newton method for distributed
  optimization.
\newblock In \emph{Advances in Neural Information Processing Systems 31}, pp.\
  2332--2342. Curran Associates, Inc., 2018{\natexlab{b}}.

\bibitem[Woodruff(2014)]{woodruff2014sketching}
Woodruff, D.~P.
\newblock Sketching as a tool for numerical linear algebra.
\newblock \emph{Foundations and Trends{\textregistered} in Theoretical Computer
  Science}, 10\penalty0 (1--2):\penalty0 1--157, 2014.

\end{thebibliography}
\bibliographystyle{icml2020}

\newpage
\onecolumn

\section{Supplementary File}
We give the proofs for the all the lemmas and theorems in the supplementary file.

\textit{Note:} A Jupyter notebook (.ipynb) containing code has been separately uploaded (also as a .py file). The code requires Python 3 to run.

\subsection{Proofs of Theorems in Section \ref{sec:ihs}}
\proof[\textbf{Proof of Theorem \ref{thm_IHS_error_decay}.}]
The update rule for distributed IHS is given as
\begin{align}
    x_{t+1} = x_t - \mu \frac{1}{q} \sum_{k=1}^q (A^TS_{t,k}^TS_{t,k}A)^{-1}A^T(Ax_t-b).
\end{align}
Let us decompose $b$ as $b=Ax^* + b^\perp$ and note that $A^T b^\perp = 0$ which gives us:
\begin{align}
    x_{t+1} = x_t - \mu \frac{1}{q} \sum_{k=1}^q (A^TS_{t,k}^TS_{t,k}A)^{-1}A^TA e_t.
\end{align}
Subtracting $x^*$ from both sides, we obtain an equation in terms of the error vector $e_t$ only:
\begin{align*}
    e_{t+1} &= e_t - \mu \frac{1}{q} \sum_{k=1}^q (A^TS_{t,k}^TS_{t,k}A)^{-1}A^TA e_t \\
    &= \left(I - \mu \frac{1}{q} \sum_{k=1}^q (A^TS_{t,k}^TS_{t,k}A)^{-1}A^TA \right) e_t.
\end{align*}
Let us multiply both sides by $A$ from the left and define $Q_{t,k} \coloneqq A(A^TS_{t,k}^TS_{t,k}A)^{-1}A^T$ and we will have the following equation:
\begin{align*}
    e^A_{t+1} &= \left(I - \mu \frac{1}{q} \sum_{k=1}^q Q_{t,k} \right) e^A_t.
\end{align*}

We now analyze the expectation of $\ell_2$ norm of $e^A_{t+1}$: 
\begin{align} \label{expected_error_expanded}
    \Exs [ ||e^A_{t+1}||_2^2 ] &= \Exs \left[ \left\Vert \frac{1}{q} \sum_{k=1}^q (I - \mu Q_{t,k}) e^A_t \right\Vert_2^2 \right] \nonumber \\
    &= \frac{1}{q^2} \Exs \left[ \sum_{k=1}^q \sum_{l=1}^q \langle (I - \mu Q_{t,k}) e^A_t, (I - \mu Q_{t,l}) e^A_t \rangle \right] \nonumber \\
    &= \frac{1}{q^2} \sum_{k=1}^q \sum_{l=1}^q \Exs \left[ \langle (I - \mu Q_{t,k}) e^A_t, (I - \mu Q_{t,l}) e^A_t \rangle \right].
\end{align}
The contribution for $k \neq l$ in the double summation of \eqref{expected_error_expanded} is equal to zero because for $k \neq l$, we have
\begin{align*}
    \Exs \left[ \langle (I - \mu Q_{t,k}) e^A_t, (I - \mu Q_{t,l}) e^A_t \rangle \right] &= \langle \Exs [ (I - \mu Q_{t,k}) e^A_t], \Exs[ (I - \mu Q_{t,l}) e^A_t] \rangle \\
    &= \langle \Exs [ (I - \mu Q_{t,k}) e^A_t], \Exs[ (I - \mu Q_{t,k}) e^A_t] \rangle \\
    &= \left\Vert \Exs [ (I - \mu Q_{t,k}) e^A_t] \right\Vert_2^2.
\end{align*}

The term in the last line above is zero for $\mu=\frac{1}{\theta_1}$:
\begin{align*}
    \Exs [ (I - \mu Q_{t,k}) e^A_t] &= \Exs [(I - \mu A(A^TS_{t,k}^TS_{t,k}A)^{-1}A^T) e^A_t ] \\
    &= (I - \mu \theta_1 A(A^TA)^{-1} A^T) e^A_t  \\
    &= (I - \mu \theta_1 UU^T)e^A_t \\
    &= (I - UU^T)e^A_t \\
    &= 0
\end{align*}
where we used $A=U\Sigma V^T$.
For the rest of the proof, we assume that we set $\mu = 1/\theta_1$. Now that we know the contribution from terms with $k \neq l$ is zero, the expansion in \eqref{expected_error_expanded} can be rewritten:
\begin{align*}
    \Exs [ ||e^A_{t+1}||_2^2 ] &= \frac{1}{q^2} \sum_{k=1}^q \Exs \left[ \langle (I - \mu Q_{t,k}) e^A_t, (I - \mu Q_{t,k}) e^A_t \rangle \right] \\
    &= \frac{1}{q^2} \sum_{k=1}^q \Exs[|| (I - \mu Q_{t,k}) e^A_t ||_2^2] \\
    &= \frac{1}{q} \Exs[|| (I - \mu Q_{t,1}) e^A_t ||_2^2] \\
    &= \frac{1}{q} \left( ||e^A_t||_2^2 + \mu^2 \Exs[ ||Q_{t,1} e^A_t||_2^2] - 2\mu (e^A_t)^T \Exs[Q_{t,1}]e^A_t \right) \\
    &= \frac{1}{q} \left( \mu^2 \Exs[ ||Q_{t,1} e^A_t||_2^2] - ||e^A_t||_2^2 \right) \\
    &= \frac{1}{q} \left( \mu^2 (e^A_t)^T \Exs[Q_{t,1}^T Q_{t,1}] e^A_t - ||e^A_t||_2^2 \right)
\end{align*}

The term $\Exs[Q_{t,1}^T Q_{t,1}]$ can be simplified using SVD decomposition $A=U\Sigma V^T$. This gives us $Q_{t,k} = U(U^TS_{t,k}^TS_{t,k}U)^{-1}U^T$ and furthermore we have:
\begin{align*}
    \Exs[Q_{t,1}^T Q_{t,1}] &= \Exs[ U(U^TS_{t,1}^TS_{t,1}U)^{-1}U^T U(U^TS_{t,1}^TS_{t,1}U)^{-1}U^T ] \\
    &=  \Exs[ U(U^TS_{t,1}^TS_{t,1}U)^{-1} (U^TS_{t,1}^TS_{t,1}U)^{-1}U^T ] \\
    &= U \Exs[(U^TS_{t,1}^TS_{t,1}U)^{-2}] U^T \\
    &= \theta_2 UU^T.
\end{align*}
Plugging this in, we obtain:
\begin{align*}
    \Exs [ ||e^A_{t+1}||_2^2 ] &= \frac{1}{q} \left( \theta_2 \mu^2 (e^A_t)^T  UU^T e^A_t - ||e^A_t||_2^2 \right) \\
    &= \frac{1}{q} \left( \theta_2 \mu^2 ||U^T e^A_t||_2^2 - ||e^A_t||_2^2 \right) \\
    &= \frac{1}{q} \left( \theta_2 \mu^2 ||e^A_t||_2^2 - ||e^A_t||_2^2 \right) \\
    &= \frac{\theta_2 \mu^2 - 1}{q}  ||e^A_t||_2^2 \\
    &= \frac{1}{q} \left( \frac{\theta_2}{\theta_1^2} - 1 \right) ||e^A_t||_2^2 
\end{align*}
\qed

\begin{proof} [\textbf{Proof of Corollary \ref{ihs_main_corollary}.}]
Taking the expectation with respect to $S_{t,k}$, $k=1,...,q$ of both sides of the equation given in Theorem \ref{thm_IHS_error_decay}, we obtain
\begin{align*}
    \Exs [||e^A_{t+1}||_2^2] = \frac{1}{q} \left(\frac{\theta_2}{\theta_1^2}-1\right) \Exs [||e^A_t||_2^2] \, .
\end{align*}
This gives us the relationship between the initial error (when we initialize $x_0$ to be the zero vector) and the expected error in iteration $t$:
\begin{align*}
    \Exs [||e^A_t||_2^2] = \frac{1}{q^t} \left(\frac{\theta_2}{\theta_1^2}-1\right)^t ||Ax^*||_2^2.
\end{align*}
It follows that the expected error reaches $\epsilon$-accuracy with respect to the initial error at iteration $T$ where:
\begin{align*}
    \frac{1}{q^T} \left(\frac{\theta_2}{\theta_1^2}-1\right)^T &= \epsilon \\
    q^T \left(\frac{\theta_2}{\theta_1^2}-1\right)^{-T} &= \frac{1}{\epsilon} \\
    T\left(\log(q)-\log\left(\frac{\theta_2}{\theta_1^2}-1\right) \right) &= \log(1/\epsilon) \\
    T &= \frac{\log(1/\epsilon)}{\log(q)-\log\left(\frac{\theta_2}{\theta_1^2}-1\right)}.
\end{align*}

Each iteration requires communicating a $d$-dimensional vector for every worker, and we have $q$ workers and the algorithm runs for $T$ iterations, hence the communication load is $Tqd$.

The computational load per worker at each iteration involves the following numbers of operations:
\begin{itemize}
    \item Sketching $A$: $mnd$ multiplications
    \item Computing $\tilde{H}_{t,k}$: $md^2$ multiplications
    \item Computing $g_t$: $\mathcal{O}(nd)$ operations
    \item Solving $\tilde{H}_{t,k}^{-1} g_t$: $\mathcal{O}(d^3)$ operations.
\end{itemize}
\end{proof}

\subsection{Proofs of Theorems in Section \ref{sec:regularized_least_squares}}
\begin{proof} [\textbf{Proof of Lemma \ref{expectation_inverse_regularization_2}.}]
In the following, we assume that we are in the regime where $n$ approaches infinity. 

The expectation term $\Exs [ (U^TS^TSU+\lambda_2 I)^{-1}]$ is equal to the identity matrix times a scalar (i.e. $cI_d$) because it is signed permutation invariant, which we show as follows. Let $P \in \mathbb{R}^{d\times d}$ be a permutation matrix and $D \in \mathbb{R}^{d\times d}$ be an invertible diagonal sign matrix ($-1$ and $+1$'s on the diagonals). A matrix $M$ is signed permutation invariant if $(DP) M (DP)^T = M$. We note that the signed permutation matrix is orthogonal: $(DP)^T(DP) = P^TD^TDP = P^TP = I_d$, which we later use. 
\begin{align*}
    (DP)\Exs_S [ (U^TS^TSU+\lambda_2 I)^{-1} ](DP)^T &= \Exs_S [ (DP)(U^TS^TSU+\lambda_2 I)^{-1} (DP)^T ] \\
    &= \Exs_S [ ((DP)^T U^TS^TSU(DP)+\lambda_2 I)^{-1} ] \\
    &= \Exs_{SUPD} [ \Exs_S [ ((DP)^T U^TS^TSU(DP)+\lambda_2 I)^{-1} | SUPD ]] \\
    &= \Exs_{SUPD} [ ((DP)^T U^TS^TSU(DP)+\lambda_2 I)^{-1} ] \\
    &= \Exs_{SU^\prime} [ ({U^\prime}^T S^TSU^\prime +\lambda_2 I)^{-1} ]
\end{align*}
where we made the variable change $U^\prime = UDP$ and note that $U^\prime$ has orthonormal columns because $DP$ is an orthogonal transformation. $SUPD$ and $SU$ have the same distribution because $PD$ is an orthogonal transformation and $S$ is a Gaussian matrix. This shows that $\Exs [ (U^TS^TSU+\lambda_2 I)^{-1}]$ is signed permutation invariant.

Now that we established that $\Exs [ (U^TS^TSU+\lambda_2 I)^{-1}]$ is equal to the identity matrix times a scalar, we move on to find the value of the scalar. We use the identity $\Exs_{DP}[(DP) Q (DP)^T] = \frac{\tr Q}{d}I_d$ for $Q\in \mathbb{R}^{d\times d}$ where the diagonal entries of $D$ are sampled from the Rademacher distribution and $P$ is sampled uniformly from the set of all possible permutation matrices. We already established that $\Exs [ (U^TS^TSU+\lambda_2 I)^{-1}]$ is equal to $(DP)\Exs_S [ (U^TS^TSU+\lambda_2 I)^{-1} ](DP)^T$ for any signed permutation matrix of the form $DP$. It follows that
\begin{align*}
    \Exs [ (U^TS^TSU+\lambda_2 I)^{-1}] &= (DP)\Exs_S [ (U^TS^TSU+\lambda_2 I)^{-1} ](DP)^T \\
    &= \frac{1}{|R|}\sum_{DP \in R} (DP)\Exs_S [ (U^TS^TSU+\lambda_2 I)^{-1} ](DP)^T \\
    &= \Exs_{DP} [ (DP)\Exs_S [ (U^TS^TSU+\lambda_2 I)^{-1} ](DP)^T ] \\
    &= \frac{1}{d} \tr (\Exs_S [ (U^TS^TSU+\lambda_2 I)^{-1} ]) I_d 
\end{align*}
where we define $R$ to be the set of all possible signed permutation matrices $DP$ in going from line 1 to line 2. 

By Lemma \ref{expectation_inverse_regularization}, the trace term is equal to $d \times \theta_3(d/m, \lambda_2)$, which concludes the proof.
\end{proof}

\begin{proof} [\textbf{Proof of Theorem \ref{opt_lambda_2_newton_LS}.}]
Closed form expressions for the optimal solution and the output of the $k$'th worker are as follows:
\begin{align*}
    x^* &= (A^TA + \lambda_1 I_d)^{-1} A^Tb, \\
    \hat{x}_k &= (A^TS_k^TS_kA + \lambda_2 I_d)^{-1}A^TS_k^TS_kb.
\end{align*}
Equivalently, $x^*$ can be written as:
\begin{align*}
    x^* = \arg\min \left\Vert \begin{bmatrix} A \\ \sqrt{\lambda_1}I_d \end{bmatrix} x - \begin{bmatrix} b \\ 0_d \end{bmatrix} \right\Vert_2^2.
\end{align*}
This allows us to decompose $\begin{bmatrix} b \\ 0_d \end{bmatrix}$ as 
\begin{align*}
    \begin{bmatrix} b \\ 0_d \end{bmatrix} = \begin{bmatrix} A \\ \sqrt{\lambda_1}I_d \end{bmatrix} x^* + b^\perp
\end{align*}
where $b^\perp = \begin{bmatrix} b_1^\perp \\ b_2^\perp \end{bmatrix}$ with $b_1^\perp \in \mathbb{R}^{n}$ and $b_2^\perp \in \mathbb{R}^d$. From the above equation we obtain $b^\perp_2 = - \sqrt{\lambda_1}x^*$ and $\begin{bmatrix} A^T & \sqrt{\lambda_1}I_d \end{bmatrix} b^\perp = A^Tb_1^\perp +\sqrt{\lambda_1}b_2^\perp = 0$. 

The bias of $\hat{x}_k$ is given by (omitting the subscript $k$ in $S_k$ for simplicity)
\begin{align} 
    &\Exs[ A (\hat{x}_k - x^*)] = \nonumber \\
    &= \Exs[A(A^TS^TSA + \lambda_2 I_d)^{-1}A^TS^TSb - Ax^*] \nonumber\\
    &= \Exs[ U(U^TS^TSU+\lambda_2 \Sigma^{-2})^{-1}U^TS^TSb] - Ax^* \nonumber\\
    &= \Exs[ U(U^TS^TSU+\lambda_2 \Sigma^{-2})^{-1}U^TS^TS(Ax^*+b_1^\perp)] - Ax^* \nonumber\\
    &= \Exs[ U(U^TS^TSU+\lambda_2 \Sigma^{-2})^{-1}U^TS^TSU\Sigma V^Tx^* + U(U^TS^TSU+\lambda_2 \Sigma^{-2})^{-1}U^TS^TSb_1^\perp] - Ax^* \nonumber\\
    &= \Exs[ U(U^TS^TSU+\lambda_2 \Sigma^{-2})^{-1}(U^TS^TSU+\lambda_2 \Sigma^{-2}-\lambda_2 \Sigma^{-2})\Sigma V^Tx^* + U(U^TS^TSU+\lambda_2 \Sigma^{-2})^{-1}U^TS^TSb_1^\perp] - Ax^* \nonumber\\
    &= \Exs[ - \lambda_2 U(U^TS^TSU+\lambda_2 \Sigma^{-2})^{-1}\Sigma^{-1}V^Tx^*] + \Exs[U(U^TS^TSU+\lambda_2 \Sigma^{-2})^{-1}U^TS^TSb_1^\perp]. \nonumber
\end{align}
By the assumption $\Sigma=\sigma I_d$, the bias becomes
\begin{align} \label{eq:bias_expect_decomp}
    \Exs[ A (\hat{x}_k - x^*)] = \Exs[ - \lambda_2 \sigma^{-1} U(U^TS^TSU+\lambda_2 \sigma^{-2} I_d)^{-1}V^Tx^*] + \Exs[U(U^TS^TSU+\lambda_2 \sigma^{-2} I_d)^{-1}U^TS^TSb_1^\perp].
\end{align}
The first expectation term of \eqref{eq:bias_expect_decomp} can be evaluated using Lemma \ref{expectation_inverse_regularization_2} (as $n$ goes to infinity):
\begin{align} \label{eq:first_expect_term}
    \Exs[ - \lambda_2 \sigma^{-1} U(U^TS^TSU+\lambda_2 I_d)^{-1}V^Tx^*] = -\lambda_2 \sigma^{-1} \theta_3(d/m, \lambda_2 \sigma^{-2}) UV^Tx^*.
\end{align}
To find the second expectation term of \eqref{eq:bias_expect_decomp}, let us first consider the full SVD of $A$ given by $A = \begin{bmatrix} U & U^\perp \end{bmatrix} \begin{bmatrix} \Sigma \\ 0_{(n-d) \times d} \end{bmatrix} V^T$ where $U \in \mathbb{R}^{n\times d}$ and $U^\perp \in \mathbb{R}^{n\times (n-d)}$. The matrix $\begin{bmatrix} U & U^\perp \end{bmatrix}$ is an orthogonal matrix, which implies $UU^T + U^\perp (U^\perp)^T = I_d$. If we insert $UU^T + U^\perp (U^\perp)^T = I_d$ between $S$ and $b_1^\perp$, the second term of \eqref{eq:bias_expect_decomp} becomes
\begin{align*}
    &\Exs[U(U^TS^TSU+\lambda_2 \sigma^{-2} I_d)^{-1}U^TS^TSb_1^\perp] = \\
    &=\Exs[U(U^TS^TSU+\lambda_2 \sigma^{-2} I_d)^{-1}U^TS^TS(UU^T + U^\perp (U^\perp)^T)b_1^\perp] \\
    &= \Exs[U(U^TS^TSU+\lambda_2 \sigma^{-2} I_d)^{-1}U^TS^TS UU^T b_1^\perp] + \Exs[U(U^TS^TSU+\lambda_2 \sigma^{-2} I_d)^{-1}U^TS^TS U^\perp (U^\perp)^T b_1^\perp] \\
    &= \Exs[U(U^TS^TSU+\lambda_2 \sigma^{-2} I_d)^{-1}U^TS^TS UU^T b_1^\perp] \\
    &= \Exs[U(U^TS^TSU+\lambda_2 \sigma^{-2} I_d)^{-1}(U^TS^TS U + \lambda_2 \sigma^{-2} I_d - \lambda_2 \sigma^{-2} I_d) U^T b_1^\perp] \\
    &= U (I_d - \lambda_2 \sigma^{-2} \Exs [ (U^TS^TSU+\lambda_2 \sigma^{-2} I_d)^{-1} ]) U^T b_1^\perp \\
    &= (1 - \lambda_2 \sigma^{-2} \theta_3(d/m, \lambda_2\sigma^{-2})) UU^T b_1^\perp
\end{align*}
where in the fourth line we have used $\Exs_S [ U(U^TS^TSU+\lambda_2 \sigma^{-2} I_d)^{-1}U^TS^TS U^\perp (U^\perp)^T b_1^\perp ] = \Exs_{SU} [\Exs_S  [ U(U^TS^TSU+\lambda_2 \sigma^{-2} I_d)^{-1}U^TS^TS U^\perp (U^\perp)^T b_1^\perp | SU]] = 0$ since $\Exs_S[SU^\perp|SU] = 0$ as $U$ and $U^\perp$ are orthogonal. The last line follows from Lemma \ref{expectation_inverse_regularization_2}, as $n$ goes to infinity.

Note that $U^T b_1^\perp = \lambda_1 \Sigma^{-1}V^Tx^*$ and for $\Sigma = \sigma I_d$, this becomes $U^T b_1^\perp = \lambda_1 \sigma^{-1} V^Tx^*$.

Bringing all of these pieces together, we have the bias equal to (as $n$ goes to infinity):
\begin{align*}
    \Exs[ A (\hat{x}_k - x^*)] &= -\lambda_2 \sigma^{-1} \theta_3(d/m, \lambda_2 \sigma^{-2}) UV^Tx^* + \lambda_1 \sigma^{-1} (1 - \lambda_2 \sigma^{-2} \theta_3(d/m, \lambda_2\sigma^{-2})) UV^Tx^* \\
    &= \sigma^{-1} (\lambda_1 - \lambda_2 \theta_3(d/m, \lambda_2 \sigma^{-2}) ( 1 + \lambda_1 \sigma^{-2})) UV^Tx^*.
\end{align*}
If there is a value of $\lambda_2 > 0$ that satisfies $\lambda_1 - \lambda_2 \theta_3(d/m, \lambda_2 \sigma^{-2}) ( 1 + \lambda_1 \sigma^{-2}) = 0$, then that value of $\lambda_2$ makes $\hat{x}_k$ an unbiased estimator. Equivalently,
\begin{align*}
    \left( \frac{-\lambda_2\sigma^{-2}+d/m-1 + \sqrt{(-\lambda_2\sigma^{-2} + d/m-1)^2+4\lambda_2\sigma^{-2}d/m}}{2\sigma^{-2} d/m} \right) &= \frac{\lambda_1}{1+\lambda_1\sigma^{-2}} \\
    -\lambda_2\sigma^{-2}+d/m-1 + \sqrt{(-\lambda_2\sigma^{-2} + d/m-1)^2+4\lambda_2\sigma^{-2}d/m} &= 2 \frac{d}{m \sigma^2} \frac{\lambda_1}{1+\lambda_1\sigma^{-2}},
\end{align*}
where we note that the LHS is a monotonically increasing function of $\lambda_2$ in the regime $\lambda_2 \geq 0$ and it attains its minimum in this regime at $\lambda_2=0$. Analyzing this equation using these observations, for the cases of $m > d$ and $m \leq d$ separately, we find that for the case of $m \leq d$, we need the following to be satisfied for zero bias:
\begin{align*}
    2\frac{d}{m\sigma^2}\frac{\lambda_1}{1+\lambda_1/\sigma^2} &\geq 2\left(\frac{d}{m}-1\right), \\
    \lambda_1 &\geq \sigma^2 \left( \frac{d}{m} - 1\right),
\end{align*}
whereas there is no condition on $\lambda_1$ for the case of $m > d$.

The value of $\lambda_2$ that will lead to zero bias is given by
\begin{align*}
    \lambda_2^* = \lambda_1 - \frac{d}{m}\frac{1}{1+\lambda_1/\sigma^2}.
\end{align*}

\end{proof}

\subsection{Proofs of Theorems in Section \ref{sec:newton_sketch}}
\begin{proof} [\textbf{Proof of Theorem \ref{thm:dist_newton_sketch}.}]
The optimal update direction is given by
\begin{align*}
    \Delta_t^* =((H_t^{1/2})^T H_t^{1/2})^{-1}g_t = H_t^{-1} g_t
\end{align*}
and the estimate update direction due to a single sketch is given by
\begin{align*}
    \hat{\Delta}_{t,k} = \alpha_s ((H_t^{1/2})^TS_{t,k}^TS_{t,k}H_t^{1/2})^{-1}g_t.
\end{align*}
where $\alpha_s \in \mathbb{R}$ is the step size scaling factor to be determined.

Letting $S_{t,k}$ be a Gaussian sketch, the bias can be written as
\begin{align*}
    \Exs[H_t^{1/2}(\hat{\Delta}_{t,k} - \Delta_t^*)] &= \Exs[\alpha_s H_t^{1/2}((H_t^{1/2})^TS_{t,k}^TS_{t,k}H_t^{1/2})^{-1}g_t - H_t^{1/2}H_t^{-1}g_t] \\
    &= \alpha_s H_t^{1/2} \Exs[((H_t^{1/2})^TS_{t,k}^TS_{t,k}H_t^{1/2})^{-1}]g_t - H_t^{1/2}H_t^{-1}g_t \\
    &= \alpha_s \theta_1 H_t^{1/2}((H_t^{1/2})^TH_t^{1/2})^{-1}g_t - H_t^{1/2}H_t^{-1}g_t \\
    &= \left( \alpha_s \theta_1 - 1 \right) H_t^{1/2}H_t^{-1}g_t.
\end{align*}
In the third line, we plug in the mean of $((H_t^{1/2})^TS_{t,k}^TS_{t,k}H_t^{1/2})^{-1}$ which is distributed as inverse Wishart distribution (see Lemma \ref{theta1_theta2_lemma_TEMP}). This calculation shows that the single sketch estimator gives an unbiased update direction for $\alpha_s = 1/\theta_1$.

The variance analysis is as follows:
\begin{align*}
    &\Exs[\|H_t^{1/2}(\hat{\Delta}_{t,k} - \Delta_t^*)\|_2^2] = \Exs[\hat{\Delta}_{t,k}^TH_t\hat{\Delta}_{t,k} + {\Delta_t^*}^TH_t\Delta_t^* - 2{\Delta_t^*}^TH_t\hat{\Delta}_{t,k}] \\
    &= \alpha_s^2 g_t^T \Exs[ ((H_t^{1/2})^TS_{t,k}^TS_{t,k}H_t^{1/2})^{-1} H_t ((H_t^{1/2})^TS_{t,k}^TS_{t,k}H_t^{1/2})^{-1} ]g_t + g_t^T H_t^{-1} g_t - 2\alpha_s g_t^T \Exs[((H_t^{1/2})^TS_{t,k}^TS_{t,k}H_t^{1/2})^{-1}] g_t \\
    &= \alpha_s^2 g_t^T \Exs[ ((H_t^{1/2})^TS_{t,k}^TS_{t,k}H_t^{1/2})^{-1} H_t ((H_t^{1/2})^TS_{t,k}^TS_{t,k}H_t^{1/2})^{-1} ]g_t + \left(1 - 2 \alpha_s \theta_1 \right) g_t^T H_t^{-1} g_t.
\end{align*}
Plugging $H_t^{1/2}=U\Sigma V^T$ into the first term and assuming $H_t^{1/2}$ has full column rank, we obtain
\begin{align*}
    g_t^T \Exs[ ((H_t^{1/2})^TS_{t,k}^TS_{t,k}H_t^{1/2})^{-1} H_t ((H_t^{1/2})^TS_{t,k}^TS_{t,k}H_t^{1/2})^{-1} ]g_t &= g_t^T V\Sigma^{-1} \Exs[ (U^TS_{t,k}^TS_{t,k}U)^{-2} ] \Sigma^{-1}V^T g_t \\
    &= g_t^T V\Sigma^{-1} (\theta_2 I_d) \Sigma^{-1}V^T g_t \\
    &= \theta_2 g_t^T V \Sigma^{-2} V^T g_t,
\end{align*}
where the second line follows due to Lemma \ref{theta1_theta2_lemma_TEMP}. Because $H_t^{-1} = V\Sigma^{-2}V^T$, the variance becomes:
\begin{align*}
    \Exs[\|H_t^{1/2}(\hat{\Delta}_{t,k} - \Delta_t^*)\|_2^2] &= \left( \alpha_s^2 \theta_2 + 1 - 2 \alpha_s \theta_1\right) g_t^T V\Sigma^{-2}V^T g_t \\
    &= \left( \alpha_s^2 \theta_2 + 1 - 2 \alpha_s \theta_1\right) \| \Sigma^{-1}V^Tg_t \|_2^2.
\end{align*}

It follows that the variance is minimized when $\alpha_s$ is chosen as $\alpha_s = \theta_1 / \theta_2$.
\end{proof}

\blems [\cite{lacotte2019faster}] \label{theta1_theta2_lemma_TEMP}
For the Gaussian sketch matrix $S \in \mathbb{R}^{m\times n}$ with i.i.d. entries distributed as $\mathcal{N}(0,1/\sqrt{m})$ where $m \geq d$, and for $U\in \mathbb{R}^{n\times d}$ with $U^TU=I_d$, the following are true:
\begin{align}
    \Exs[(U^TS^TSU)^{-1}] &= \theta_1 I_d, \nonumber \\
    \Exs[(U^TS^TSU)^{-2}] &= \theta_2 I_d,
\end{align}
where $\theta_1$ and $\theta_2$ are defined as
\begin{align} \label{eq:theta_1_2_definitions_TEMP}
    \theta_1 &\coloneqq \frac{m}{m-d-1}, \nonumber \\
    \theta_2 &\coloneqq \frac{m^2(m-1)}{(m-d)(m-d-1)(m-d-3)}.
\end{align}

\elems

\subsection{Proofs of Theorems in Section \ref{sec:regularized_newton}}

\begin{proof} [\textbf{Proof of Theorem \ref{opt_lambda_2_newton}.}]
In the following, we omit the subscripts in $S_{t,k}$ for simplicity. Using the SVD decomposition of $H_t^{1/2} = U\Sigma V^T$, the bias can be written as
\begin{align*}
    \Exs[ H_t^{1/2}(\hat{\Delta}_{t,k} - \Delta_t^*)] &= U\Exs[ (U^TS^TSU + \lambda_2 \Sigma^{-2})^{-1}]\Sigma^{-1}V^Tg_t - U (I_d + \lambda_1 \Sigma^{-2})^{-1} \Sigma^{-1}V^Tg_t.
\end{align*}
By the assumption that $\Sigma = \sigma I_d$, the bias simplifies to
\begin{align*}
    \Exs[ H_t^{1/2}(\hat{\Delta}_{t,k} - \Delta_t^*)] &= \sigma^{-1} U\Exs[ (U^TS^TSU + \lambda_2 \sigma^{-2} I_d)^{-1}]V^Tg_t - \sigma^{-1} U (I_d + \lambda_1 \sigma^{-2} I_d)^{-1} V^Tg_t \\
    &= \sigma^{-1} U\Exs[ (U^TS^TSU + \lambda_2 \sigma^{-2} I_d)^{-1}]V^Tg_t - \sigma^{-1} (1+\lambda_1\sigma^{-2})^{-1} U V^Tg_t.
\end{align*}
By Lemma \ref{expectation_inverse_regularization_2}, as $n$ goes to infinity, we have
\begin{align*}
    \Exs[ H_t^{1/2}(\hat{\Delta}_{t,k} - \Delta_t^*)] &= \sigma^{-1} \left( \theta_3(d/m, \lambda_2\sigma^{-2}) - \frac{1}{1+\lambda_1\sigma^{-2}} \right) UV^Tg_t \\
    &= \sigma^{-1} \left( \frac{-\lambda_2\sigma^{-2}+d/m-1 + \sqrt{(-\lambda_2\sigma^{-2} + d/m-1)^2+4\lambda_2\sigma^{-2}d/m}}{2\lambda_2\sigma^{-2} d/m} - \frac{1}{1+\lambda_1\sigma^{-2}} \right) UV^Tg_t.
\end{align*}
The bias becomes zero for the value of $\lambda_2$ that satisfies the following equation:
\begin{align} \label{eq:lhs_defn}
    \frac{-\lambda_2\sigma^{-2}+d/m-1 + \sqrt{(-\lambda_2\sigma^{-2} + d/m-1)^2+4\lambda_2\sigma^{-2}d/m}}{2\lambda_2\sigma^{-1} d/m} &= \frac{1}{1+\lambda_1\sigma^{-2}} \nonumber \\
    -\sigma^{-2}+\frac{1}{\lambda_2} \left(\frac{d}{m}-1\right) +  \sqrt{\left(-\sigma^{-2} + \frac{1}{\lambda_2}\left(\frac{d}{m}-1 \right) \right)^2+4\sigma^{-2}\frac{d}{m\lambda_2}} &= 2\sigma^{-1}\frac{d}{m} \frac{1}{1+\lambda_1\sigma^{-2}}.
\end{align}
In the regime where $\lambda_2 \geq 0$, the LHS of \eqref{eq:lhs_defn} is always non-negative and is monotonically decreasing in $\lambda_2$.The LHS approaches zero as $\lambda_2 \rightarrow \infty$. We now consider the following cases:
\begin{itemize}
    \item Case 1: $m \leq d$. Because $d/m-1 \geq 0$, as $\lambda_2 \rightarrow 0$, the LHS goes to infinity. Since the LHS can take any values between $0$ and $\infty$, there is an appropriate $\lambda_2^*$ value that makes the bias zero for any $\lambda_1 \geq 0$ value.

    \item Case 2: $m > d$. In this case, $d/m-1<0$. The maximum of LHS in this case is reached as $\lambda_2 \rightarrow 0$ and it is equal to $2\sigma^{-2}\frac{d}{m-d}$. 
    As long as $2\sigma^{-1}\frac{d}{m} \frac{1}{1+\lambda_1\sigma^{-2}} \leq 2\sigma^{-2}\frac{d}{m-d}$ is true, then we can drive the bias down to zero. More simply, this corresponds to $\lambda_1\sigma^{-2} \geq -d/m$, which is always true because $\lambda_1 \geq 0$ and $\sigma \geq 0$. Therefore in the case of $m > d$ as well, there is a $\lambda_2^*$ value for any $\lambda_1 \geq 0$ that will drive the bias down to zero.
\end{itemize}
To sum up, for any given non-negative $\lambda_1$ value, it is possible to find a $\lambda_2^*$ value to make the sketched update direction unbiased. The optimal value for $\lambda_2$ is given by $LHS^{-1}(2\sigma^{-1}\frac{d}{m} \frac{1}{1+\lambda_1\sigma^{-2}})$, which, after some simple manipulation steps, is found to be:
\begin{align*}
    \lambda_2^* = \frac{\lambda_1+\sigma^2\frac{d}{m}}{1 + \frac{d}{m} \frac{1}{1+(\lambda_1/\sigma^2)}}.
\end{align*}
\end{proof}






\end{document}